\documentclass[10pt,twocolumn,letterpaper]{article}

\usepackage[pagenumbers]{cvpr}

\definecolor{cvprblue}{rgb}{0.21,0.49,0.74}
\usepackage[pagebackref,breaklinks,colorlinks,allcolors=cvprblue]{hyperref}

\usepackage{multirow}
\usepackage{multicol}
\usepackage{pifont}
\usepackage{amsmath} 
\usepackage{algorithm,algorithmic}
\usepackage{xcolor}
\usepackage{listings}
\makeatletter
\AfterEndEnvironment{algorithm}{\let\@algcomment\relax}
\AtEndEnvironment{algorithm}{\kern2pt\hrule\relax\vskip3pt\@algcomment}
\let\@algcomment\relax
\newcommand\algcomment[1]{\def\@algcomment{\footnotesize#1}}
\renewcommand\fs@ruled{\def\@fs@cfont{\bfseries}\let\@fs@capt\floatc@ruled
  \def\@fs@pre{\hrule height.8pt depth0pt \kern2pt}%
  \def\@fs@post{}%
  \def\@fs@mid{\kern2pt\hrule\kern2pt}%
  \let\@fs@iftopcapt\iftrue}
\makeatother

\newcommand{\gray}[1]{\textcolor{gray}{#1}}
\newcommand{\green}[1]{\textcolor[RGB]{96,177,87}{#1}}
\newcommand{\fn}[1]{\footnotesize{#1}}
\newcommand{\gbf}[1]{\green{\bf{\fn{(#1)}}}}
\newcommand{\rbf}[1]{\gray{\bf{\fn{(#1)}}}}

\title{Continual SFT Matches Multimodal RLHF with Negative Supervision}

\author{
	Ke Zhu$^{1,2*}$ \quad
	Yu Wang$^2$\thanks{Equal Contributions} \quad
	Yanpeng Sun$^{2,3}$ \quad
	Qiang Chen$^2$ \quad
    Jiangjiang Liu$^2$ \quad \\
    Gang Zhang$^2$ \quad
    Jingdong Wang$^2$\thanks{Corresponding Author} \\
$^1$Nanjing University \quad
$^2$Baidu VIS \\
$^3$Nanjing University of Science and Technology \\
{\tt\small zhuk@lamda.nju.edu.cn, \{wangyu106,wangjingdong\}@baidu.com}
}

\begin{document}
\maketitle

\begin{abstract}
Multimodal RLHF usually happens after supervised finetuning (SFT) stage to continually improve vision-language models' (VLMs) comprehension. Conventional wisdom holds its superiority over continual SFT during this preference alignment stage. In this paper, we observe that the inherent value of multimodal RLHF lies in its negative supervision, the logit of the rejected responses. We thus propose a novel negative supervised finetuning (nSFT) approach that fully excavates these information resided. Our nSFT disentangles this negative supervision in RLHF paradigm, and continually aligns VLMs with a simple SFT loss. This is more memory efficient than multimodal RLHF where 2 (e.g., DPO) or 4 (e.g., PPO) large VLMs are strictly required. The effectiveness of nSFT is rigorously proved by comparing it with various multimodal RLHF approaches, across different dataset sources, base VLMs and evaluation metrics. Besides, fruitful of ablations are provided to support our hypothesis. We hope this paper will stimulate further research to properly align large vision language models.
\end{abstract}

\section{Introduction}
\label{sec:intro}

Large vision-language models (VLMs) emerged~\cite{LLM_Llava,LLM_MiniGPT4} thanks to the intelligence of large language models (LLMs). Typically, such models~\cite{LLM_BLIP2,LLaVa1.5,intervl1.5} usually experience a pretraning stage with diverse image-text pairs before supervised finetuned (SFT) in a multitask fashion~\cite{sft-nlp-multitask,sft-nlp-less-more,LLaVa1.5}. 

The successful application of reinforcement learning from human feedback (RLHF, also called preference alignment) in LLM~\cite{DPO,PPO} has shed light on vision-language areas~\cite{VLLM_DPO,seva,bpo,LLaVa-RLHF}, which aims to further leverage VLMs' multimodal comprehension ability after the standard SFT stage and to reduce potential image hallucinations. 

\begin{figure*}
	\centering

    \begin{subfigure}{0.3\linewidth}
		\includegraphics[width=0.95\linewidth]{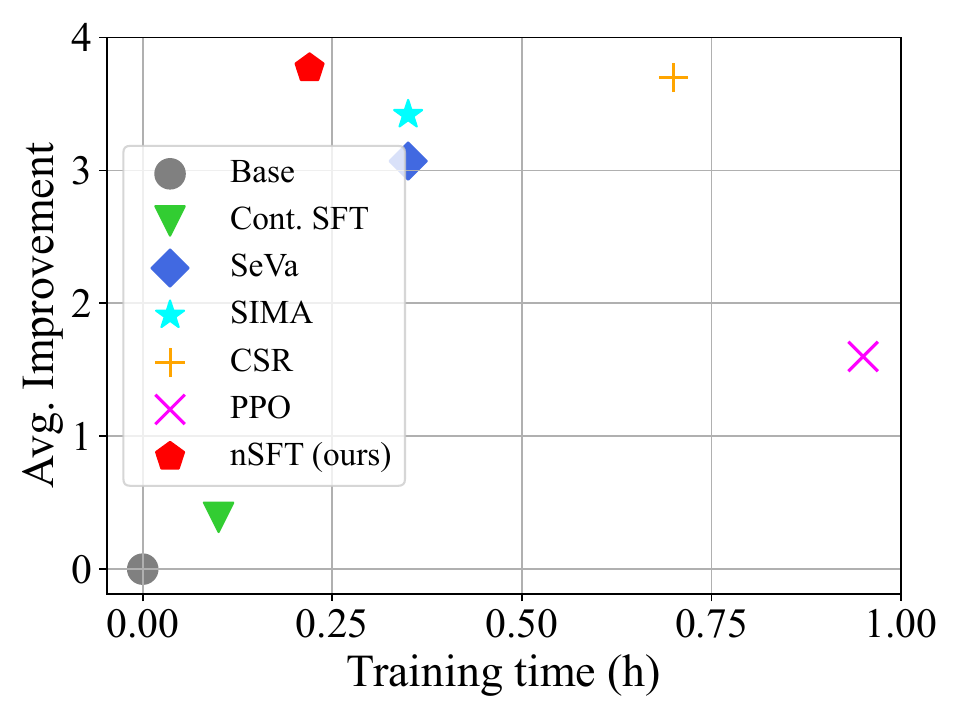}
		\caption{Training time}
    	\label{fig:method-train-time}
	\end{subfigure}
    \begin{subfigure}{0.3\linewidth}
		\includegraphics[width=0.95\linewidth]{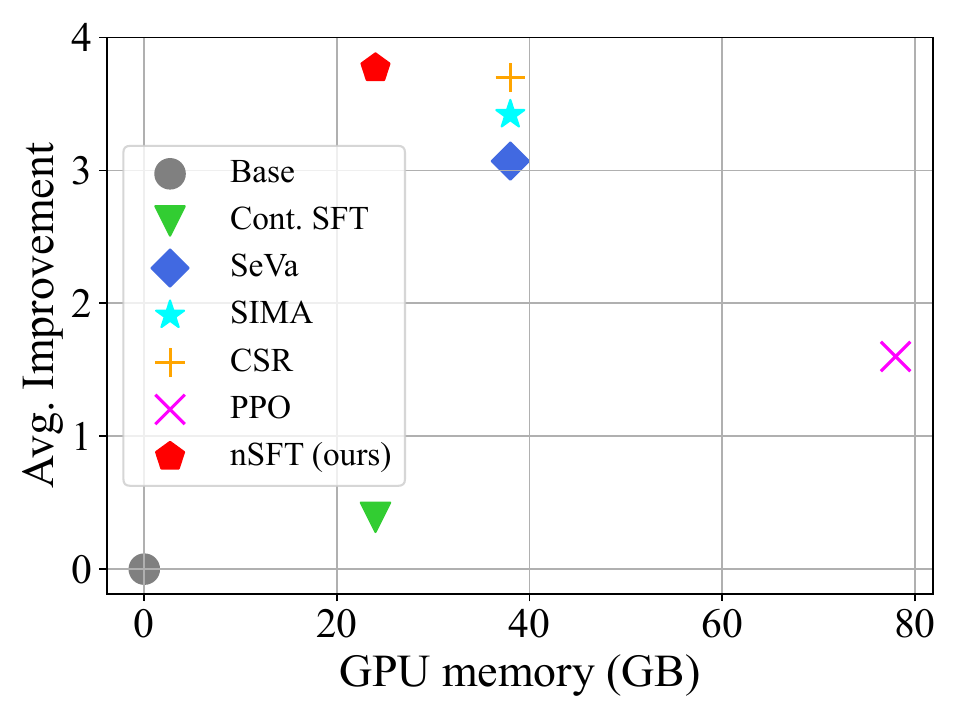}
		\caption{GPU memory consumption}
    	\label{fig:method-gpu-memory}
	\end{subfigure}
    \begin{subfigure}{0.32\linewidth}
		\includegraphics[width=0.99\linewidth]{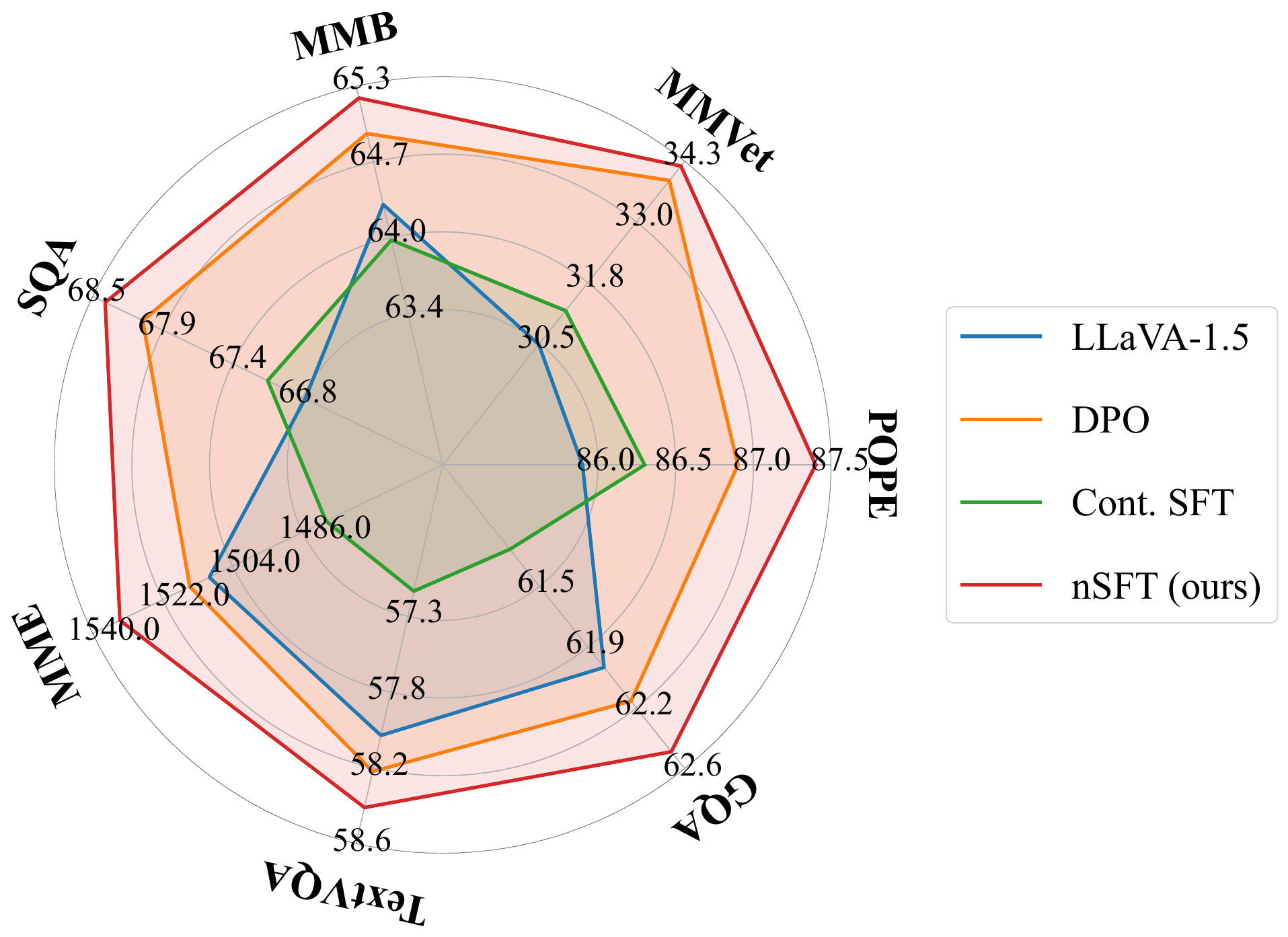}
		\caption{Multimodal comprehension tasks}
    	\label{fig:method-circle}
	\end{subfigure}

    \caption{\ref{fig:method-train-time}-\ref{fig:method-gpu-memory} shows training time, GPU memory of LLaVa-1.5 (`Base'), 3 DPO techniques: SeVa~\cite{seva}, SIMA~\cite{sima} and CSR~\cite{csr}, 1 PPO technique~\cite{LLaVa-RLHF}. \ref{fig:method-circle} shows multimodal results of averaged DPO methods, pure continual (Cont.) SFT and our nSFT.}
\end{figure*}

A widespread belief in multimodal RLHF is they all assume the \emph{inferiority} of SFT during this preference alignment process~\cite{self-play,seva,bpo}. They evidently shows that continual SFT (e.g., applying SFT in preference alignment) falls short in addressing image-level hallucination~\cite{seva} and are becoming less effective in data utilization~\cite{self-play,bpo}. These opinions holds \emph{true}, as also demonstrated in our quantitative results in Fig.~\ref{fig:method-circle} (\cf `Cont. SFT' vs `DPO'). However, we found practically that multimodal RLHF is \emph{not} as perfect as it does, since it will face GPU memory shortage~\cite{seva} and usually confronts unstable training issue~\cite{LLaVa-RLHF}.

\begin{figure}
	\centering
    \includegraphics[width=0.99\linewidth]{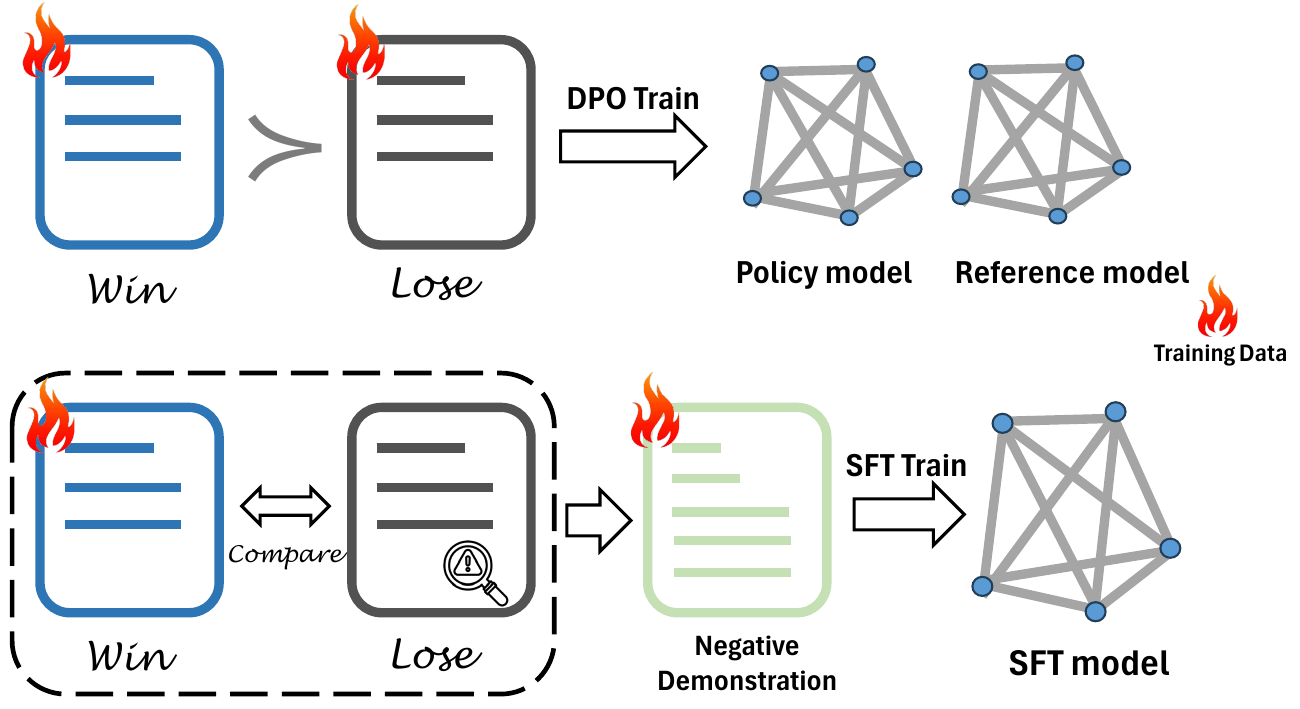}
    \caption{Standard DPO training pipeline (in the first row) and our proposed nSFT (in the second row).}
	\label{fig:method-letter}
\end{figure}

In this paper, we aim to address the following question: \emph{is multimodal RLHF indeed superior than continual SFT during preference alignment?} Through theoretical analysis in gradient and optimization aspects, we find key success of multimodal RLHF (e.g., DPO~\cite{DPO}) mostly attributes to the \emph{negative supervision} in the rejected responses, whereas \emph{naive} continual SFT fails to capture.

\begin{figure*}
	\centering
	\includegraphics[width=0.99\linewidth]{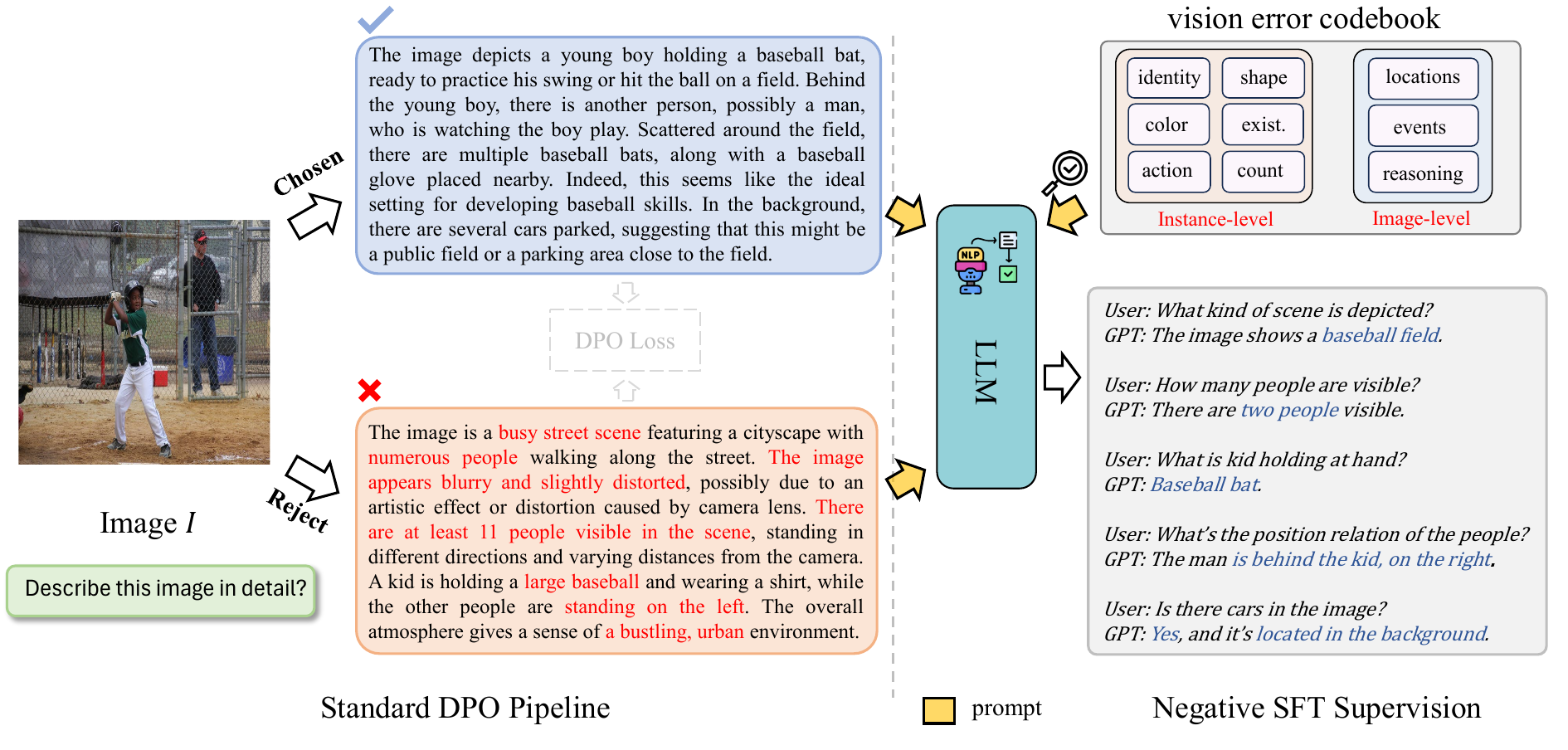}
    	\caption{Visualization of a standard DPO pipeline and our nSFT method. In DPO, GT annotations usually directly serves as chosen responses~\cite{bpo,self-play}. During nSFT, we ask an LLM to first identifies the specific error (red) part in the rejected response, by referring to chosen responses and the numerated error in the vision error codebook (\cf appendix). The LLM then constructs a conversation talking about this image that help the model avoid making such mistakes (e.g., correct answers are blue coded). This figure is best viewed in color.}
	\label{fig:method-main}
\end{figure*}

Based on this discovery, this paper proposes a novel \emph{negative supervied finetuning} (nSFT) pipeline that continually aligns VLMs with a simple SFT loss. Specifically, an LLM (e.g., GPT-4) is adopted to identify the misaligned details in the negative supervision, by referring to a large vision error codebook that contains both instance- and image-level error. The LLM then constructs a new conversation talking about the image, helping the model to realize these self mistakes from these rejected responses (\cf Fig.~\ref{fig:method-main}).

Our proposed approach disentangles the negative supervision from RLHF optimization. This makes our training more efficient in both memory and training time, as multimodal RLHF usually requires 2 or 4 models during alignment stage, as shown in Fig.~\ref{fig:method-gpu-memory}.

To verify whether nSFT matches multimodal RLHF with this \emph{constructed} negative supervision, we carefully select different preference alignment data, covering different image types and responses length. Then, a controlled experiment is done with nSFT, SFT and 3 multimodal RLHF (GT-DPO, SeVa~\cite{seva}, SIMA~\cite{sima}) methods. Experiment in Table~\ref{tab:main-improvement} shows that nSFT not only remedy pure SFT, but also achieves the overall best result across evaluation metrics. Our hypothesis also holds for more powerful VLMs like LLaVA-NeXT~\cite{llava-next}, and variants of multimodal RLHF approaches (PPO~\cite{PPO} and iterative DPO~\cite{csr}), too.

Finally, our ablations shows that nSFT can be further improved with a similar KL constraint adopted in RLHF~\cite{ChatGPT}, and captures more fine-grained detail in comparison. Our study reveals that multimodal RLHF is still more effective in addressing worst cases in comparison (e.g., reduce the frequency of the worst response sentence or tokens). 

Overall, our contributions are as follows:
\begin{itemize}
    \item We analyze key factor (\emph{negative supervision}) that makes multimodal RLHF successful, and propose a novel nSFT method that fully excavate this negative supervision.
    \item Our nSFT strictly matches multimodal preference alignment methods (both DPO and PPO) under different training database, scale and numerous evaluation metrics.
    \item We provide fruitful ablations to verify the generalization ability of nSFT as well as common RLHF approaches, stimulating future research to properly align large VLMs.
\end{itemize}

\section{Related Work}
\textbf{Supervised finetuning (SFT).} The notion of instruction tuning (also called supervised finetuning) initially derives from natural language processing (NLP) domain~\cite{sft-nlp-multitask,sft-nlp-scale-instruct}, with the aim to fully unlock the models response intelligence~\cite{sft-nlp-less-more} with diverse formatted prompts~\cite{sft-nlp-finetuned-llm}. In vision-language area, the SFT paradigm is similar, given that the VLMs are \emph{sufficiently} pretrained with abundant image-text pairs~\cite{ShareGPT4V,DreamLLM,chatspot}. Such technique basically aims to conduct image-based multi-task training~\cite{vary} that these models can behave well in diverse zero-shot benchmarks~\cite{arcana,LLM_MiniGPT4}. For examples, VLMs will demonstrate strong multimodal comprehension skills after this stage, such as multi-turn conversation, math and image reasoning~\cite{LLaVa1.5,cogvlm}. A distinct feature of SFT is that only positive targeted are settled, without letting the model know its \emph{wrong answers}. In this paper, we found that this property of SFT might hinder model from further improvement.

\textbf{Multimodal preference alignment.} Multimodal preference alignment~\cite{dpo-or-ppo,HA-DPO,vDPO_Gemini} (also called RLHF) happens after SFT stage to continual align VLMs with user intentions and to reduce hallucinations. Direct preference optimization~\cite{DPO,merge-optimizer-dpo,sft-implicit-regular-dpo} (DPO), one representative in this family, becomes the most popular due to its flexibility and efficiency.~\cite{self-play,bpo} adopt groudtruth (GT) as chosen response and self-response as negative, which iteratively improve the model in a GT-DPO style. More recent work, such as SeVa~\cite{seva}, adopts noised model output as rejected samples to conduct preference optimization. Effective but, we have observed a shared point in these works~\cite{seva,self-play,bpo}: they all assume the inferiority of continual SFT during preference alignment stage. In this paper, we discover the root cause of such phenomenon: \emph{negative supervision}. We advocate that continual SFT could match the performance of multimodal RLHF with this supervision signal fully integrated.

\section{Method}
\label{sec:method}
We will first introduce background of multimodal SFT and RLHF, then discuss their loss function relations. Based on this, we propose a negative supervised finetuning (nSFT) approach that continually improve VLMs' capability.

\subsection{Preliminaries}
\textbf{SFT.} Most VLMs are trained with a next token prediction loss (e.g., SFT loss). Specifically, for a given image input $I$, it went through an $H$ (combination of vision encoder and connector~\cite{mls}) to get the latent embeddings $\boldsymbol{v}$: $\boldsymbol{v} = H(I)$, which are concatenated with the question embeddings $q$: $\boldsymbol{x} = (\boldsymbol{v}, q)\,$, and are fed into a large language model $\pi_{\theta}(\cdot)$ that sequentially produces the next token:
\begin{equation}
    \pi_{\theta}(\boldsymbol{y}|\boldsymbol{x}) = \prod_{i=1}^{L} \pi_{\theta}(y_i|y_{<i}, \boldsymbol{x})\,.
    \label{eq:sft-sequential}
\end{equation}
Here $L$ denotes the response token length. During SFT, a cross entropy loss is applied to Eq.~\ref{eq:sft-sequential}, as follows:
\begin{align}
    \mathcal{L_\text{sft}} (\boldsymbol{y}) = -\sum_{i=1}^{L} \log \pi_{\theta}(y_i | y_{<i}, \boldsymbol{x}) \,. \label{eq:sft-loss-2} 
\end{align}

\textbf{RLHF and DPO.} RLHF aims to further align LLMs or VLMs with user specific intentions. In RLHF, a reward model $r_{\xi}(\cdot)$ is parameterized by $\xi$ and usually obtained by optimizing a Bradley-Terry (BT) model~\cite{DPO}:
\begin{equation}
    p^*(\boldsymbol{y}_{c} \succ \boldsymbol{y}_{r} |\boldsymbol{x}) = \frac{\exp{(r^*(\boldsymbol{x}, \boldsymbol{y_c}))}}{\exp{(r^*(\boldsymbol{x}, \boldsymbol{y_c}))} + \exp{(r^*(\boldsymbol{x}, \boldsymbol{y_r}))}}\,.
    \label{eq:bt-model}
\end{equation}
In Eq.~\ref{eq:bt-model}, $r^*(\cdot)$ is the optimal reward model, and $\boldsymbol{y}_c, \boldsymbol{y}_{r}$ represent the chosen and rejected response, respectively. $r_{\xi}(\cdot)$ are usually obtained through MLE~\cite{DPO} using preference dataset $D$ containing set of $(\boldsymbol{x}, \boldsymbol{y}_c, \boldsymbol{y}_r)$. After that, we would obtain the policy model $\pi_{\theta'}$ by maximizing:
\begin{equation}
    \max_{\theta'}\mathbb{E}_{\boldsymbol{x},\boldsymbol{y}}\left\{ r_{\xi}(\boldsymbol{x}, \boldsymbol{y}) - \beta \mathbb{D}_\text{KL}[\pi_{\theta'}(\boldsymbol{y}|\boldsymbol{x}) | \pi_{\text{ref}}(\boldsymbol{y}|\boldsymbol{x}) ] \right\}\,,
    \label{eq:rlhf}
\end{equation}
where $\pi_{\theta'},\pi_{\text{ref}}$ represents the policy model and reference model, respectively. $\mathbb{D}_\text{KL}$ and $\beta$ means the KL divergence and the hyper-parameter. DPO simplifies the optimization process of RLHF by resorting to the closed form of Eq.~\ref{eq:rlhf}:
\begin{equation}
    r^*(x, \boldsymbol{y}) = \beta \log\frac{\pi^*(\boldsymbol{y}|\boldsymbol{x})}{\pi_{\text{ref}}(\boldsymbol{y}|\boldsymbol{x})} + C\,,
\label{eq:closed-form}
\end{equation}
in which the optimal reward model $r^*$ is a function of the optimal policy model $\pi^*$ ($C$ is a constant). Thus, we can directly optimize the BT model in Eq.~\ref{eq:bt-model} by substituting Eq.~\ref{eq:closed-form} into it, and get the final DPO loss function:
\begin{equation}
    \mathcal{L}_{\text{d}} = -\mathbb{E}_{D}\left[\log \sigma\left(\beta \log\frac{\pi_{\theta'}(\boldsymbol{y}_c|\boldsymbol{x})}{\pi_{\text{ref}}(\boldsymbol{y}_c|\boldsymbol{x})}- \beta\log\frac{\pi_{\theta'}(\boldsymbol{y}_r|\boldsymbol{x})}{\pi_{\text{ref}}(\boldsymbol{y}_r|\boldsymbol{x})}\right)\right]
    \label{eq:dpo}
\end{equation}
In Eq.~\ref{eq:dpo}, $\sigma(\cdot)$ denotes the sigmoid function that normalize the DPO logtis, and the whole optimization process is conducted by sampling preference tuple $(\boldsymbol{x}, \boldsymbol{y}_c, \boldsymbol{y}_r)$ from $D$.

\subsection{Negative supervision matters}
\label{sec:3-2}
We will take DPO for illustrations, given its wide popularity in VLMs~\cite{bpo,seva,sima}. Note we also quantitatively verify PPO in Table~\ref{tab:ablate-nsft-ppo} to make our hypothesis more general.

\textbf{Relations between DPO and SFT.} Following many previous multimodal RLHF methods~\cite{self-play,bpo}, we take GT-DPO to illustrate their relations (e.g., $\boldsymbol{y}_c,\boldsymbol{y}_r$ are GT annotations and model self response, respectively). From Eq.~\ref{eq:dpo}, we can define the `logit' in DPO loss function as:
\begin{equation}
    p_{\text{dpo}} =  \log\frac{\pi_{\theta'}(\boldsymbol{y}_c|\boldsymbol{x})}{\pi_{\text{ref}}(\boldsymbol{y}_c|\boldsymbol{x})}- \log\frac{\pi_{\theta'}(\boldsymbol{y}_r|\boldsymbol{x})}{\pi_{\text{ref}}(\boldsymbol{y}_r|\boldsymbol{x})}\,.
    \label{eq:dpo-logit}
\end{equation}
If we omit the reference constraint in Eq.~\ref{eq:dpo-logit} and sequentially stretch the productive function of $\pi$, we will obtain:
\begin{equation}
\begin{aligned}
    p'_{\text{dpo}} &= \sum_{i} \left[\log \pi_{\theta'} (y_{c, i}|y_{c, <i}, \boldsymbol{x}) -  \log \pi_{\theta'} (y_{r, i}|y_{r, <i}, \boldsymbol{x}) \right] \\
    &= - (\mathcal{L}_{\text{sft}} (\boldsymbol{y}_c) - \mathcal{L}_{\text{sft}}(\boldsymbol{y}_r))
    \label{eq:dpo-sft}
\end{aligned}
\end{equation}
That is, the core component of DPO is \emph{actually} the subtraction of two SFT loss: the chosen and reject sequences. With more strict derivation (\cf appendix), we show DPO's gradient is a linear combination of two SFT gradient:
\begin{equation}
    \frac{\partial \mathcal{L}_{d}}{\partial \theta'} =\frac{1}{p_\text{dpo}}\left[ \frac{\partial \mathcal{L}_{\text{sft}}(\boldsymbol{y}_c)}{\partial \theta'} - \frac{\partial \mathcal{L}_{\text{sft}}(\boldsymbol{y}_r)}{\partial \theta'} \right] \,,
\end{equation}
and the only missing information comes from the absence of rejected SFT loss $\mathcal{L}_{\text{sft}}(\boldsymbol{y}_r)$. We thus conclude: \emph{the inferior performance mainly derive from the lack of negative supervision resided in the rejected responses!} 

\textbf{The role of negative supervision.} We further investigate the importance of negative supervision. Note that a con-current work in NLP discover a similar trend~\cite{limitation-dpo}, but not aiming at relating to SFT. Here we first simplify Eq.~\ref{eq:dpo}:
\begin{equation}
    t_1 = \frac{\pi_{\theta'}(\boldsymbol{y}_c|\boldsymbol{x})}{\pi_{\text{ref}}(\boldsymbol{y}_c|\boldsymbol{x})}, \quad t_2 = \frac{\pi_{\theta'}(\boldsymbol{y}_r|\boldsymbol{x})}{\pi_{\text{ref}}(\boldsymbol{y}_r|\boldsymbol{x})}\,,
\end{equation}
and reformulate the DPO loss function as follows:
\begin{align}
    \mathcal{L}_d &= - \log (\frac{t_1^{\beta}}{t_1^{\beta} + t_2^{\beta}})\,,
\end{align}
where we safely ignore the expectation term $\mathbb{E}_{D}$ for better clarification. Then the gradient with regard to $t_1,t_2$ is:
\begin{equation}
\frac{\partial \mathcal{L}}{\partial t_1} =  \frac{-\beta t_2^{\beta}}{t_1 (t_1^{\beta} + t_2^{\beta})},\,\,\,\, \frac{\partial \mathcal{L}}{\partial t_2} =  \frac{\beta t_2^{\beta - 1}}{t_1 (t_1^{\beta} + t_2^{\beta})}\,. 
\label{eq:dpo-partial}
\end{equation}
And we can obtain the update rate ratio as:
\begin{equation}
    \left|\frac{\partial \mathcal{L}}{\partial t_1} / \frac{\partial \mathcal{L}}{\partial t_2} \right| = t_2 / t_1 \,.
\end{equation}
Inspired by the conclusion in~\cite{limitation-dpo}, we know for any regular preference pair, $t_2 / t_1 < 1$ holds, such that the optimization will be biased towards how to reject samples, with faster gradient updating rate of $t_1$ (\cf appendix and~\cite{limitation-dpo}).

As continual SFT is already lacking negative supervision, this optimization bias could exaggerate this effect by leaning gradient more towards rejecting samples, such that continual SFT could probably lag behind. This hypothesis was then firmly verified by quantitative experiments in Fig.~\ref{fig:method-circle}, where pure continual SFT is much inferior than DPO (using a \emph{same} training dataset). These analysis calls for a new SFT method that could alleviate this issue.

\subsection{Our nSFT approach} 
\label{sec:3-3}

\textbf{Disentangle negative supervision.} Since the negative supervision is deeply entangled in DPO `logit' and \emph{pairwise} relations exists, it is hard to obtain those negative supervision by directly optimizing in an \emph{SFT style} as Eq.~\ref{eq:sft-loss-2}. As a result, we involve a construction function $G(\cdot)$ (an LLM like GPT-4), and define our loss as (note that $\boldsymbol{y}_c$ equals the GT captions, as we are discussing in GT-DPO phase):
\begin{equation}
    \mathcal{L}_{\text{nSFT}} = \mathcal{L}_{\text{sft}} (\boldsymbol{y}_c)  + \mathcal{L}_{\text{sft}}(G(\boldsymbol{y}_r ))\,.
\end{equation}
The aim of $G(\cdot)$ is identify and re-organize the false information embedded in the rejected responses $\boldsymbol{y}_r$, where the model can learn from in an SFT manner.

\begin{table*}
	\setlength{\tabcolsep}{5pt}

	\footnotesize
	\centering
	\begin{tabular}{lllllllllllllll}
		\toprule[1pt]
		   \multicolumn{1}{c|}{\multirow{2}{*}{Alignment Data}} & \multicolumn{1}{c|}{\multirow{2}{*}{Method}}  & \multicolumn{4}{c|}{Traditional VQA}& \multicolumn{4}{c|}{MM Comprehension} & \multicolumn{4}{c}{Hallucinations}\\

      & \multicolumn{1}{|c}{} & \multicolumn{1}{|c}{SQA}  & GQA & VQA$^{\text{T}}$ &  \multicolumn{1}{c|}{total} & MMVet & MME & MMB  &\multicolumn{1}{c|}{total}  & POPE & CHAIR $\downarrow$ & MMHal & \multicolumn{1}{c}{total}\\
		\midrule[1pt]

    \multirow{6}{*}{OCRVQA~\cite{OCRVQA}} 
        & baseline & 66.8 & 62.0 & 58.0 & \gbf{+0.0} & 30.5 & 1510 & 64.3 & \gbf{+0.0} & 85.9 & 32.0 & 2.80 & \gbf{+0.0}  \\
        & GT-DPO & 67.8 & 61.4 & 57.7 & \gbf{+0.1} & 32.5 & 1412 & 63.9 & \gbf{+1.6} & 84.3 & 31.5 & 2.90 & \gbf{+0.6}  \\
        & SeVa & 67.6 & 62.0 & 57.5 & \gbf{+0.3} & 32.5 & 1502 & 64.9 & \gbf{+2.6} & 86.6 & 27.3 & 3.00 & \gbf{+8.7}  \\
        & SIMA & 68.0 & 61.9 & 58.2 & \gbf{+1.3} & 32.5 & 1486 & 64.8 & \gbf{+2.5} & 86.2 & 29.4 & 2.93 & \gbf{+5.1}  \\
        & Cont. SFT & 67.9 & 61.7 & 56.9 & \rbf{-0.3} & 33.3 & 1490 & 64.5 & \gbf{+3.0} & 87.0 & 34.0 & 2.76 & \rbf{-1.6}  \\
        & nSFT & \textbf{68.1} & \textbf{62.0} & \textbf{58.1} & \gbf{+1.4} & \textbf{34.0} & \textbf{1515} & \textbf{64.9} & \gbf{+4.1} & \textbf{87.1} & \textbf{26.5} & \textbf{2.93} & \gbf{+8.9}  \\

    \midrule[1pt]

    \multirow{6}{*}{TextCaps~\cite{TextVQA}} 
        & baseline & 66.8 & 62.0 & 58.0 & \gbf{+0.0} & 30.5 & 1510 & 64.3 & \gbf{+0.0} & 85.9 & 32.0 & 2.80 & \gbf{+0.0} \\
        & GT-DPO & 68.0 & 61.7 & 57.5 & \gbf{+0.4} & 34.2 & 1500 & 64.2 & \gbf{+3.6} & 86.5 & 29.2 & 2.83 & \gbf{+3.9} \\
        & SeVa & 68.1 & 61.7 & 57.8 & \gbf{+0.8} & 34.6 & 1480 & 65.0 & \gbf{+4.8} & 86.3 & 26.3 & 2.90 & \gbf{+7.8} \\
        & SIMA & 68.0 & 62.1 & 58.0 & \gbf{+1.3} & 32.2 & 1473 & 64.9 & \gbf{+2.3} & 85.9 & 27.6 & 2.87 & \gbf{+5.6} \\
        & Cont. SFT & 66.9 & 61.3 & 56.6 & \rbf{-2.0} & 31.0 & 1520 & 64.4 & \gbf{+0.6} & 86.3 & 30.5 & 2.83 & \gbf{+2.4} \\
        & nSFT & \textbf{68.4} & \textbf{62.3} & \textbf{58.2} & \gbf{+2.1} & \textbf{33.7} & \textbf{1521} & \textbf{65.3} & \gbf{+4.2} & \textbf{87.2} & \textbf{26.2} & \textbf{2.97} & \gbf{+9.9} \\

    \midrule[1pt]

    \multirow{6}{*}{LLaVA-150k~\cite{LLaVa1.5}} 
        & baseline & 66.8 & 62.0 & 58.0 & \gbf{+0.0} & 30.5 & 1510 & 64.3 & \gbf{+0.0} & 85.9 & 32.0 & 2.80 & \gbf{+0.0} \\
        & GT-DPO & 68.1 & 61.6 & 57.6 & \gbf{+0.5} & 33.9 & 1497 & 63.9 & \gbf{+3.0} & 85.9 & 30.7 & 2.80 & \gbf{+1.3} \\
        & SeVa & 67.5 & 61.4 & 58.0 & \gbf{+0.1} & 32.5 & 1490 & 64.7 & \gbf{+2.4} & 85.6 & 28.2 & 2.94 & \gbf{+5.8} \\
        & SIMA & 67.9 & 62.2 & 58.2 & \gbf{+1.5} & 32.1 & 1511 & 64.9 & \gbf{+2.2} & 86.9 & 26.2 & 2.97 & \gbf{+9.6} \\
        & Cont. SFT & 67.1 & 60.9 & 57.0 & \rbf{-1.8} & 31.2 & 1480 & 64.0 & \gbf{+0.4} & 86.3 & 29.1 & 2.91 & \gbf{+5.1} \\
        & nSFT & \textbf{68.4} & \textbf{62.3} & \textbf{58.4} & \gbf{+2.3} & \textbf{34.2} & \textbf{1550} & \textbf{65.2} & \gbf{+4.6} & \textbf{87.4} & \textbf{25.4} & \textbf{3.02} & \gbf{+11.8} \\
  
		\bottomrule[1pt]
	\end{tabular}
\caption{Nine benchmark results by applying 5 continual learning methods with 3 training data sources. We categorize these benchmarks into 3 categories, namely traditional VQA, multimodal (MM) comprehension and hallucination. We also show total score (`total') by summing up each column's improvement. `CHAIR' refers to average score of `CHAIR$^\text{i}$' and `CHAIR$^\text{s}$'~\cite{chair}. We adopted ChatGPT to evaluate `MMHal' score, and scale them to range 0-100 before calculating the `total'. The training data are set as 10k for each (\cf appendix for more result). Due to a mismatch evaluation caused by transformers~\cite{llava-next}, we fixed transformers at version 4.31.0~\cite{transformers}.}
\label{tab:main-improvement}
\end{table*}

\begin{figure*}
	\centering
    \begin{subfigure}{0.24\linewidth}
		\includegraphics[width=0.95\linewidth]{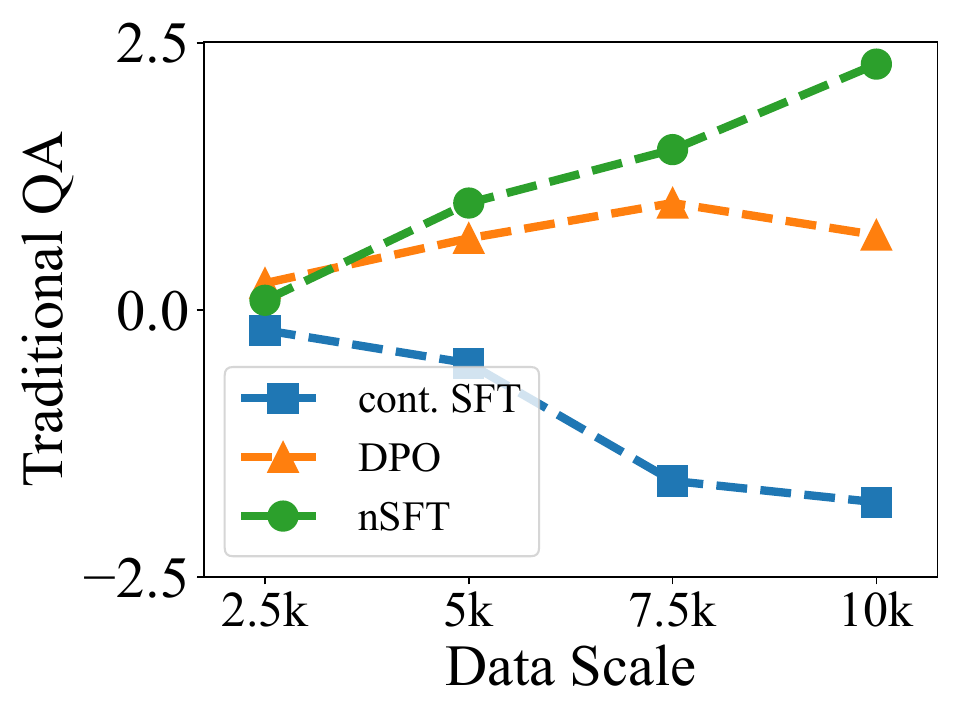}
		\caption{Traditional VQA}
		\label{fig:main-improve:vqa}
	\end{subfigure}
	\begin{subfigure}{0.24\linewidth}
		\includegraphics[width=0.95\linewidth]{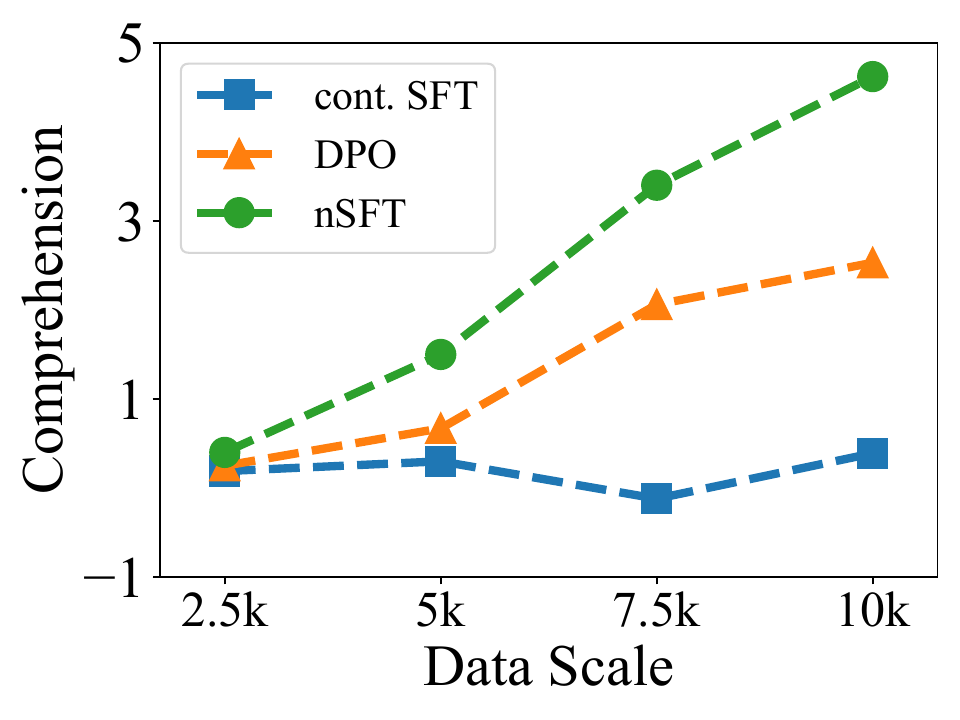}
    	\caption{Multimodal comprehension}
		\label{fig:main-improve:comprehension}
    \end{subfigure} 
	\begin{subfigure}{0.24\linewidth}
		\includegraphics[width=0.95\linewidth]{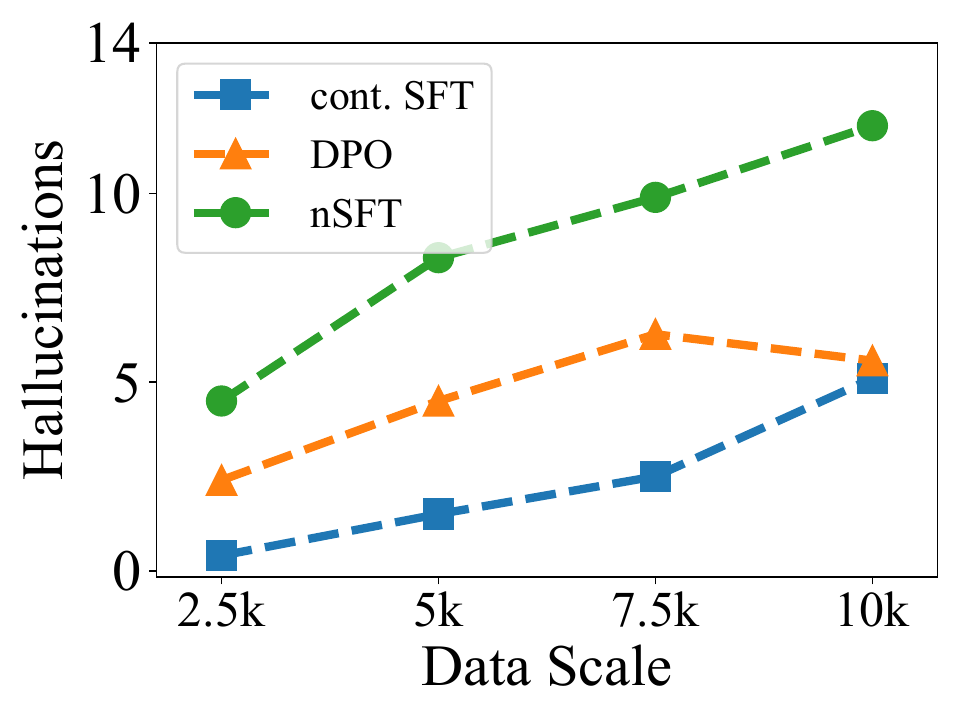}
    	\caption{Multimodal hallucination}
		\label{fig:main-improve:hallucination}
    \end{subfigure} 
	\begin{subfigure}{0.24\linewidth}
		\includegraphics[width=0.95\linewidth]{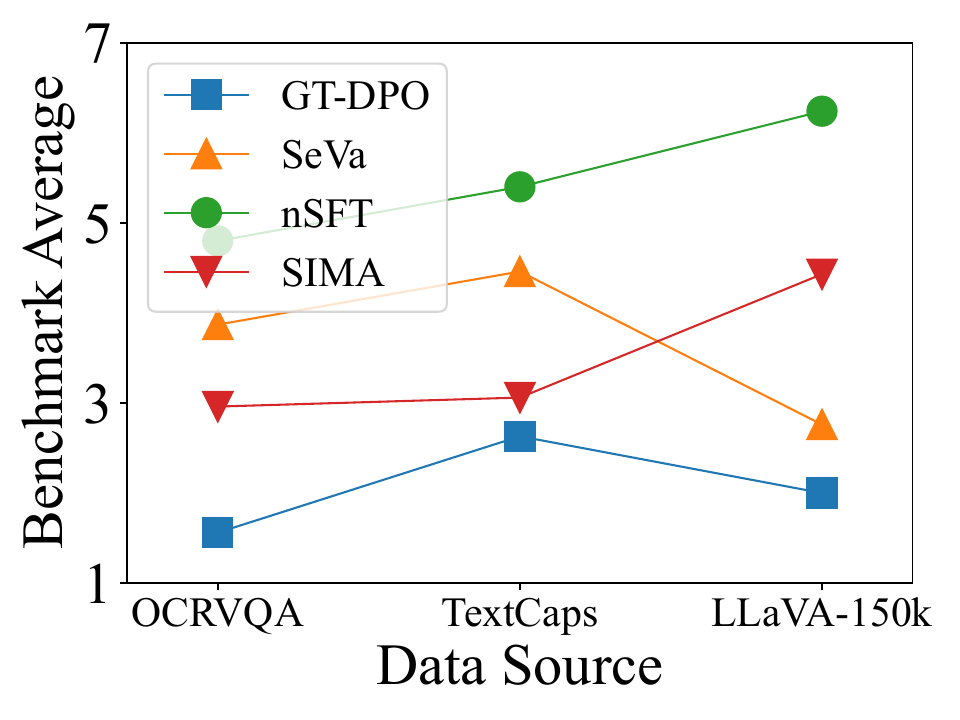}
    	\caption{Effect of dataset}
		\label{fig:main-improve:dataset}
    \end{subfigure}

	\caption{Visualization of benchmarks results (relative improvement over baseline LLaVA-1.5-7B) using continual SFT, DPO and our nSFT (\cf.~\ref{fig:main-improve:vqa}-\ref{fig:main-improve:hallucination}). Here we choose LLaVA-150k as the data source and randomly choose 2.5k, 5k, 7.5k and 10k for model alignment. In~\ref{fig:main-improve:dataset}, we visualize the averaged results (over 9 benchmarks shown in Table~\ref{tab:main-improvement}), and analyzing the effect of dataset choices.}
	\label{fig:main-improvement}
\end{figure*}

\textbf{Vision error codebook.} As the rejected responses contain abundant (possibly hallucinatory) information of the image, the construction function $G(\cdot)$ should cover as much image related error as possible. Thus, we introduce a large vision error codebook $Q$, containing all possible recognition error types, during identification. 

Specifically, an LLM is first obliged to identify both image-level and instance-level error in the rejected sentences, by referring to GT information $\boldsymbol{y}_c$ and the table $Q$. The LLM is then asked to formulate a new conversation talking about this image, to reinforce the mistakes that the model already made. That is, $\boldsymbol{y}_c, \boldsymbol{y}_r$ and $Q$ are all prompts that are fed into the LLM $G(\cdot)$ (\cf appendix). Putting it all together, we formulate our final loss of nSFT:
\begin{equation}
    \mathcal{L}_{\text{nSFT}} = \mathcal{L}_{\text{sft}} (\boldsymbol{y}_c)  + \mathcal{L}_{\text{sft}}(G(\boldsymbol{y}_r ;\boldsymbol{y}_c, Q))\,.
    \label{eq:nsft-final}
\end{equation}

 The merit of our nSFT is in two fold. On one hand, we break the pairwise relation of chosen and response $\boldsymbol{y}_c,\boldsymbol{y}_r$, as \emph{strictly} required during RLHF~\cite{seva,sima} optimization. On the other hand, the whole preference alignment is optimized with an SFT loss, much more memory efficient than multimodal RLHF where \emph{at least} 2 models are indispensible.

Note there is no KL constraint in Eq.~\ref{eq:nsft-final} as adopted in RLHF. In ablations, we show that adding a per-token constraint to nSFT will further improve the results.

\section{Experiments}
We will first introduce the specific negative construction pipeline before showing the general comparisons. Finally, fruitful of ablations are provided.

\subsection{Constructing nSFT} 
Due to a huge discrepancy of mutlimodal RLHF literature~\cite{seva,bpo} in dataset choices, we select three alignment dataset from various source: OCRVQA~\cite{OCRVQA}, TextCaps~\cite{TextVQA} and LLaVA-150k~\cite{LLaVa1.5}, covering object centric and scene images, short, medium and long responses.

\textbf{OCRVQA.} This dataset are set of object centric images containing books' title pages. During experiment, we choose 10k such data and generate model's response to construct the rejected answers, which are then directly compared (w/o an LLM) with the GT output to find out mistakes. For each mistake, we construct doubled question-answer pairs. Below shows a constructed sample where the models misclassify a travel book as a recipe book:

\texttt{Q1}: \texttt{Is this a travel book?}  \texttt{A1}: \texttt{Yes}

\texttt{Q2}: \texttt{Is this a recipe book?} \texttt{A2}: \texttt{No}

These newly constructed question-answer pair are \emph{appended} to the end of original GT conversations.

\begin{table*}
	\setlength{\tabcolsep}{5pt}
	\small
	\centering
\begin{tabular}{llllllllllll}
\toprule
     Method & Language model & VQA$^{\text{T}}$ & SQA & GQA & MMB &  MMB$^\text{CN}$ & MME & POPE & SEED$^{I}$ & SHR $(\downarrow)$ & MMVet \\
    \midrule[1pt]
    BLIP-2 & FLAN-T5 & 61.0 & – & 41.0 & – & – & 1293 & 85.3 & 46.4 & – & 22.4  \\
    InstructBLIP & Vicuna-7B & 50.1 & 60.5 & – & 36.0 & 23.7 & – & 53.4 & - & 26.2 &   \\
    InstructBLIP & Vicuna-13B & 50.7 & 63.1 & 49.5 & – & – & 1213 & 78.9 & – & 51.2 & 25.6  \\ 
    Shikra & Vicuna-13B & – & – & – & 58.8 & – & – & – & – & – & –  \\ 
    IDEFICS-9B & LLaMA-7B & – & – & 38.4 & 48.2 & 25.2 & – & – & – & – & –  \\ 
    IDEFICS-80B & LLaMA-65B & 54.8 & – & 45.2 & 54.5 & 38.1 & 1177 & – & – & – & –  \\ 
    LLaVA & Vicuna-7B & – & 38.5 & – & 34.1 & 14.1 & 807 & – & 25.5 & – & 26.7  \\ 
    LLaVA-1.5 & Vicuna-7B & 58.2 & 66.8 & 62.0 & 64.3 & 58.3 & 1510 & 85.9 & 65.7 & 36.7 & 30.5  \\ 
    \midrule
    SeVa-7B & Vicuna-7B & 56.2 & 67.5 & 60.7 & 65.6 & 59.2 & 1450 & 86.7 & 65.8 & 34.9 & \textbf{36.8}  \\ 
    SIMA-7B & Vicuna-7B & 58.3 & 68.1 & 62.2 & 64.9 & 59.0 & 1507 & 86.5 & 65.9 & 34.5 & 32.6  \\ 
    nSFT (\emph{ours}) & Vicuna-7B & \textbf{58.7} & \textbf{68.5} & \textbf{62.9} & \textbf{67.1} & \textbf{61.0} & \textbf{1531} & \textbf{86.8} & \textbf{66.2} & \textbf{34.2} & \underline{34.0}  \\ 
    \bottomrule[1pt]
\end{tabular}

\caption{Comparing our nSFT with state-of-the-art methods. We compare nSFT with various base VLM models and preference alignment methods (e.g., SeVa~\cite{seva} and SIMA~\cite{sima}). Here we adopted a mixture of OCRVQA, TextVQA and LLaVA-150k, forming a total of 15k SFT data (refer to our experimental settings for more details). For SeVa-7B and SIMA-7B, we reproduce their results using official code implementation, and evaluate them all using a same transformer version (e.g., version 4.31.0).}
    \label{tab:compare-sota}
\end{table*}

\begin{table*}
	\setlength{\tabcolsep}{4pt}
	\footnotesize
	\centering
	\begin{tabular}{l|llll|llll|llll}
		\toprule[1pt]
		\multicolumn{1}{c|}{\multirow{2}{*}{Method}}  & \multicolumn{4}{c|}{OCRVQA} & \multicolumn{4}{c|}{TextCaps} & \multicolumn{4}{c}{LLaVA-150k} \\
		& IF score & Accuracy & ACC$_{10}^{b}$ & ACC$_{10}^{w}$ &  IF score & Accuracy & ACC$_{10}^{b}$ & ACC$_{10}^{w}$ & IF score & Accuracy & ACC$_{10}^{b}$ & ACC$_{10}^{w}$ \\
		\midrule[1pt] 
        LLaVA-1.5 & 8.52 & 8.37 & 9.80 & 0.00 & 6.41 & 5.88 & 9.32 & 0.40 & 7.79 & 7.01 & 9.25 & 2.02\\
        +SeVa & 8.69 & 8.51 & 10.0 & \textbf{0.56} & 6.71 & 6.07 & 9.40 & \textbf{1.72} & 7.99 & 7.13 & 9.20 & 2.32\\
        +GT-DPO & 8.70 & 8.41 & 10.0 & 0.26 & 6.65 &  \textbf{6.22} & 9.30 & 1.13 & 7.95 & 7.12 & 9.38 & \textbf{2.68}\\
        +nSFT(\emph{ours}) & \textbf{8.71} & \textbf{8.56} & \textbf{10.0} & 0.37 & \textbf{6.78} & 6.12 & \textbf{9.50} & 0.70 & \textbf{8.09} & \textbf{7.27} & \textbf{9.62} & 2.30 \\

        \midrule[1pt]
	\end{tabular}
\caption{In-domain evaluation results. We apply SeVa, GT-DPO and our nSFT using a subset of OCRVQA, TextCaps and LLaVA-150k, respectively, and evaluate each model on \emph{its own validation data source} in an held-out manner (\cf appendix). GPT-4 is involved to judge the model's instruction following ability (`IF score') and its image relevant accuracy (`Accuracy'), ranged from 0-10. We also list the average of best/worst 10 accuracy score in `Accuracy', shown as `ACC$^b_{10}$' and `ACC$^w_{10}$', respectively.}
\label{tab:in-domain}
\end{table*}

\textbf{TextCaps \& LLaVA-150k} These databases contain image captions of medium token length (around 10 tokens per caption in TextCaps) and long captions (around 100 token in LLaVA-150k). In implementation, we choose 10k data each to generate response and compare it with groundtruth using LLM (e.g., GPT-4). The LLM will identify any hallucination in the model original output by referring to the vision error codebook and GT captions, and correspondingly construct rectified conversation. If no error exists, the LLM simply construct a conversation based on GT information. The constructed data are concatenated with original GT to form the total question-answer pairs for each image.

\subsection{Experimental settings}
\textbf{Training.} In our main Table~\ref{tab:main-improvement}, we choose LLaVA-1.5-7B as baseline and reproduce 3 different multimodal RLHF methods: GT-DPO~\cite{sima}, SeVa~\cite{seva}, SIMA~\cite{sima}, 2 SFT methods: pure continual SFT and our nSFT. We strictly reproduce SIMA and SeVa following its official implementation, and conduct GT-DPO with the \emph{same} experiment setting as SeVa~\cite{seva}. The architecture of continual SFT and nSFT all follow LLaVA-1.5~\cite{LLaVa1.5}. Specifically, during nSFT, we adopt deepspeed framework and ZeRO-3~\cite{LLaVa1.5} optimization. The batch size, learning rate and weight decay are set as 128, 2e-6 and 0, respectively, following a cosine scheduler. When conducting state-of-the-art comparison Table~\ref{tab:compare-sota}, we use a mixture of 5k OCRVQA, 5k TextCaps and 5k LLaVA-150k, and reproduce other multimodal RLHF approaches with their \emph{original data} (e.g., SIMA uses a 17k training data). We also evaluate nSFT on larger VLMs like LLaVA-1.5-13B and LLaVA-NeXT-13B~\cite{llava-next} in Table~\ref{tab:ablate-different-models}.

\textbf{Evaluation.} We mainly choose 9 benchmarks for evaluation, and categorize them into 3 main clusters, containing traditional VQA: SQA~\cite{sqa}, GQA~\cite{gqa}, TextVQA~\cite{TextVQA} (e.g. VQA$^{\text{T}}$), multimodal comprehension: MMVet~\cite{MMVet}, MME~\cite{seva}, MMB~\cite{MMBench}, and multimodal hallucination: POPE~\cite{POPE}, CHAIR~\cite{sima}, MMHal~\cite{sima}. Please refer to appendix for more discussion abount MMHal.

\subsection{General comparisons} We first compare nSFT with other RLHF methods. As shown in Table~\ref{tab:main-improvement}, all preference alignment methods lead to a general positive effect. However, pure continual SFT is shown to hurt original model's capability, or make the improvement trivial. When the negative supervision is integrated (our nSFT), the performance have seen a great boost, surpassing all preference alignment methods across benchmarks. Interestingly, a much more significant increase is observed in hallucination benchmarks in nSFT (e.g., a 15.1 increase using LLaVA-150k data). Since we involve the vision error codebook (\cf Table~\ref{tab:ablate-nsft-component}), the model would be more sensitive to image-related details afterward (\cf Fig.~\ref{fig:ablate-main-vis}). DPO paradigm is also effective, but relatively coarse grained, as shown in later visualizations in Fig.~\ref{fig:ablate-main-vis}.

We also visualize these results in Fig.~\ref{fig:main-improvement}. In Fig.~\ref{fig:main-improve:vqa}-\ref{fig:main-improve:hallucination}, the alignment data is a 10k subset of LLaVA-150k, and `DPO' is the average score of SeVa, SIMA and GT-DPO. In Fig.~\ref{fig:main-improve:dataset}, we list all continual learning method, and average their 9 benchmark results (\cf Table~\ref{tab:main-improvement}). As shown in the figure, our nSFT shows robust scaling behavior for all benchmarks, overtaking DPO and pure continual SFT under all circumstances. Besides, as the GT token length increases, the improvement of nSFT steadily increase (\cf Fig.~\ref{fig:main-improve:dataset}), while other DPO methods fluctuates. We deduce that more GT information helps the LLM to refer and be more aware of the image content, thus facilitating the learning process.

\begin{table}
	\small
	\centering
	\begin{tabular}{ccllccccc}
		\toprule[1pt]
		\multirow{2}{*}{Method}  &\multicolumn{4}{c}{Benchmark Results} \\
         & MMB & SQA  & MME & POPE \\
        \midrule
            
            LLaVa-1.5-13B  &  67.7 & 71.6 & 1531 & 85.9\\
            + cont. SFT & 66.3 & 70.8 & 1496 & 86.1\\
            + DPO & 68.4 & 71.8 & 1520 & 87.5  \\
            + nSFT (\emph{ours}) & \textbf{68.5} & \textbf{71.9} & \textbf{1551} & \textbf{88.6}\\   
        \midrule
            LLaVA-NeXT-13B  & 70.0 & 73.6 & 1565 & 86.2\\
            + cont. SFT & 68.8 & 71.2 & 1518 & 86.3\\
            + DPO & \textbf{70.2} & 74.0 & 1550 & 87.1\\
            + nSFT (\emph{ours}) & 69.7 & \textbf{74.4}  & \textbf{1570} & \textbf{87.3} \\    
        \bottomrule[1pt]
	\end{tabular}
\caption{The generalization ability of nSFT to larger, stronger VLMs. We select a subset of LLaVA-150k for continual learning.}
\label{tab:ablate-different-models}
\end{table}

\begin{table}
	\small
	\centering
	\begin{tabular}{ccllccccc}
		\toprule[1pt]
		\multirow{2}{*}{Method}  &\multicolumn{4}{c}{Benchmark Results} \\
         & MMB & SQA  & MME & POPE \\
        \midrule
            
            baseline  &  64.3 & 66.8 & 1515 & 85.9\\
            + RLHF~\cite{LLaVa-RLHF} (PPO) & 64.7 & 67.8 & 1508 & 86.2\\
            + CSR (iter 1) &  64.3\ & 68.3 & 1501 & 86.8\\
            + CSR (iter 2) & 64.4 & 68.2 & 1518 & 86.9\\
            + CSR (iter 3) & 64.4 & \textbf{68.4} & 1523 & 87.2\\
            + nSFT (\emph{ours}) & \textbf{65.2} & \textbf{68.4} & \textbf{1550} & \textbf{87.4}\\
              
        \bottomrule[1pt]
	\end{tabular}
\caption{Compare with PPO~\cite{LLaVa-RLHF} and iterative DPO method CSR~\cite{csr}. The training dataset all uses a subset of LLaVA-150k.}
\label{tab:ablate-nsft-ppo}
\end{table}

\subsection{Comparing with state-of-the-art} 

We move onto compare our nSFT with current VLMs. In Table~\ref{tab:compare-sota}, we use a mixture of 15k dataset for nSFT, comprised of 5k data for each specific data source (OCRVQA, TextCaps and LLaVA-150k). To improve the stability for nSFT, we merge this 15k mixed data into LLaVA 665k, which are jointly trained during LLaVA-1.5 SFT stage. This could be viewed as a special case of continual learning to avoid model distribution shift~\cite{Qwen}. As shown in the table, all continual alignment/learning methods leverage the baseline LLaVA-1.5-7B results, showing that there is still room for improvement even after the standard SFT stage. Our nSFT consistently surpasses other alignment methods for 9 out of the 10 benchmarks, especially on MMB. SeVa obtains the best results in MM-Vet, possibly due to its enlarged response length~\cite{seva} that is favored by GPT-4 evaluation~\cite{MMVet}.

\subsection{Ablations}
Last, we conduct a series of ablations with step-by-step controllable analysis, to reveal the relationships between SFT \& RLHF and verify the generalization ability of our nSFT.

\begin{figure}
	\centering
    \begin{subfigure}{0.49\linewidth}
		\includegraphics[width=0.99\linewidth]{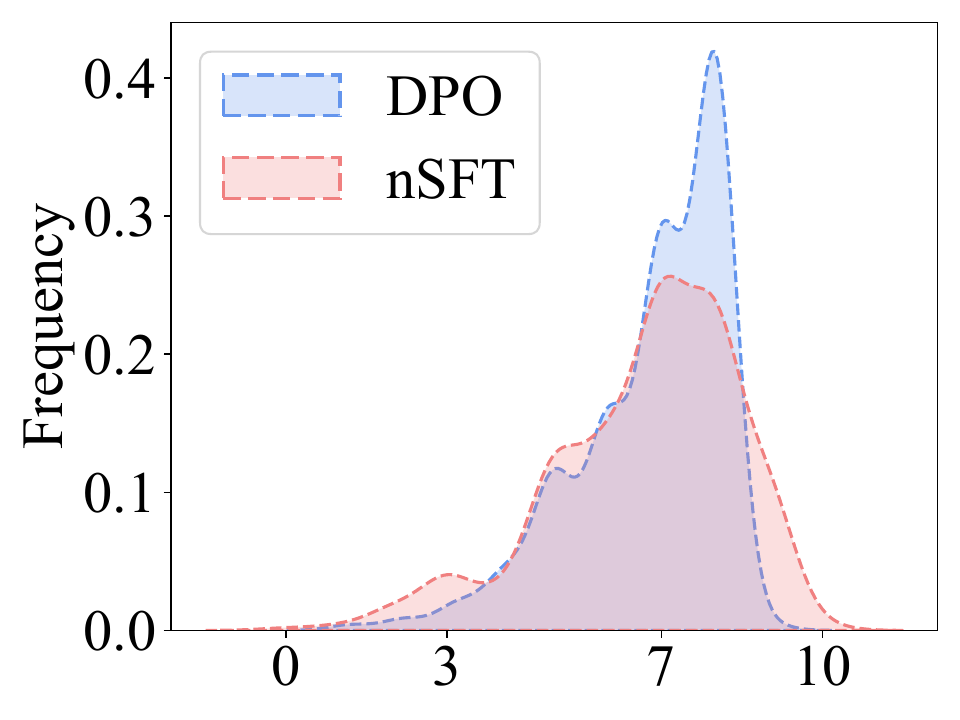}
		\caption{Instruct following score}
		\label{fig:ablate-lower-upper-if}
	\end{subfigure}
	\begin{subfigure}{0.49\linewidth}
		\includegraphics[width=0.99\linewidth]{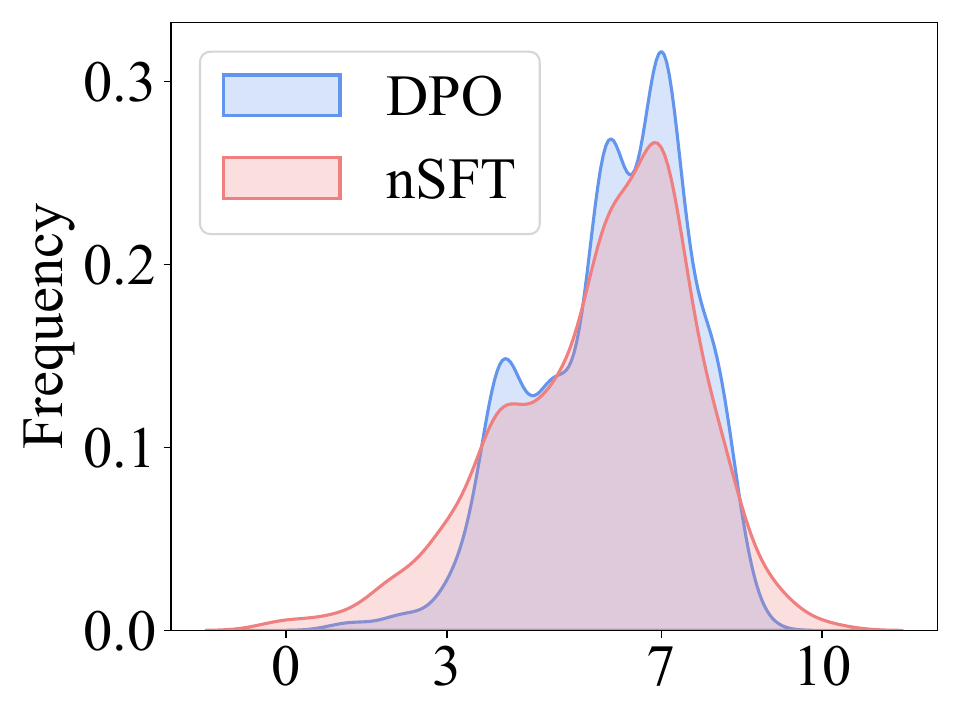}
    	\caption{Detail accuracy score}
		\label{fig:ablate-lower-upper-detail}
    \end{subfigure} 
    \caption{Visualizations of the instruct following score and detail accuracy score in TextCaps~\cite{TextVQA} dataset. The statistics are the same with Table~\ref{tab:in-domain}, and `DPO' refers to SeVa.}
	\label{fig:ablate-lower-upper}
\end{figure}

\textbf{Difference between SFT and RLHF.} We take DPO for illustration since all recent publicity~\cite{seva,sima} adopt it for its efficacy. First, an in-domain evaluation was conducted, where the LLaVA-1.5 are trained using a 10k subset of OCRVQA, TextCaps and LLaVA-150k, which are then evaluated on its own held-out data source. Not surprising in Table~\ref{tab:in-domain}, all continual learning methods improve the `IF' and `Accuracy' metrics. Besides, the best cases of our nSFT is generally higher than DPO methods, while $ACC_{10}^{w}$ leans towards DPO paradigm. Together with the previous observation in Sec.~\ref{sec:method}, we deduce that \emph{DPO mostly try to reject those worst cases, and that SFT hinges on leverage the model's best results}. Our conjecture was further verified with visualizations in Fig.~\ref{fig:ablate-lower-upper}, where a same phenomenon can be observed in the whole score distribution. 

\begin{table}
	\small
	\centering
	\begin{tabular}{ccllccccc}
		\toprule[1pt]
		\multirow{2}{*}{Method}  &\multicolumn{4}{c}{Benchmark Results} \\
         & MMB & SQA  & MME & POPE \\
        \midrule
            
            baseline  &  64.3 & 66.8 & 1515 & 85.9\\
            + nSFT &  \textbf{65.0} & \textbf{68.2} & \textbf{1533} & \textbf{86.5} \\ 
            + nSFT w/o VEC & 64.4 & 67.6 & 1505 & 86.0\\
            + nSFT w/o chosen &  64.9 & \textbf{68.2} & 1523 & 86.4 \\
              
        \bottomrule[1pt]
	\end{tabular}
\caption{The component analysis for our nSFT: the \textbf{v}ision \textbf{e}rror \textbf{c}odebook (VEC), and the involvement of chosen response during nSFT. Experiment were conducted with a subset of LLaVA-150k.}
\label{tab:ablate-nsft-component}
\end{table}

\begin{table}
	\small
	\centering
	\begin{tabular}{ccllcccc}
		\toprule[1pt]
		\multirow{2}{*}{Method} &  \multirow{2}{*}{w/ KL} &\multicolumn{3}{c}{Benchmark Results} \\
            & & MMB & VQA$^\text{T}$  & SQA \\
        \midrule
            baseline & N/A &  64.3 & 58.2 & 66.8\\
            cont. SFT & \ding{55} & 63.6 & 56.6 & 67.3 \\
            cont. SFT & \ding{51} & 64.2 & 57.8 & 67.7\\
            nSFT(\emph{ours}) & \ding{55} & 64.8 & 58.1 & 68.1 \\
            nSFT(\emph{ours}) & \ding{51} & \textbf{65.2} & \textbf{58.4} & \textbf{68.4}\\
            DPO & \ding{51}  & 65.0 & 58.1 & 68.0\\
                        
        \bottomrule[1pt]
	\end{tabular}
\caption{The effect of adding a per token KL constraint~\cite{token-dpo} in continual SFT. The DPO refers to method of SeVa.}
\label{tab:ablate-kl}
\end{table}

\begin{figure*}
	\centering
    \includegraphics[width=0.99\linewidth]{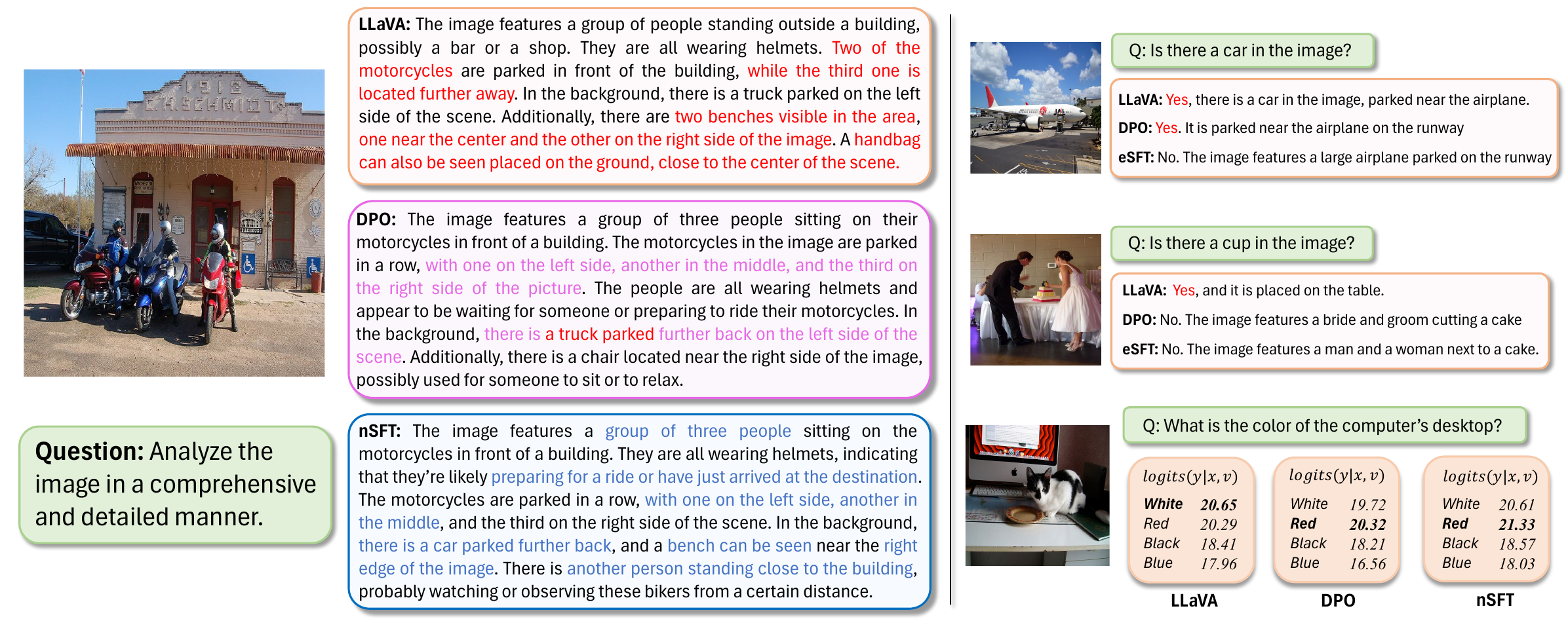}
    \caption{Visualizations of response generated from LLaVA-v1.5-7B, DPO and nSFT models. In the left part, the correct content are emphasized with purple (for DPO) and blue (for nSFT), while error content are highlighted with red. This figure is best viewed in color.}
	\label{fig:ablate-main-vis}
\end{figure*}

We also visualize the logit in Fig.~\ref{fig:ablate-main-vis} (in the lower right). Results show that DPO helps correct the errors by decreasing the logit of bad predictions (e.g., the logit of color `White'), while nSFT enhance the model by lifting those `good' logits. By referring to left of Fig.~\ref{fig:ablate-main-vis}, we found that our nSFT focus more on fine-grained details (such as counting, object, etc.), thanks to a large vision error codebook during the negative data construction. In contrast, DPO are optimized in a accumulated logit manner, and can fail to recognize such level correctness (also pointed out in~\cite{limitation-dpo}).

\textbf{Larger, stronger VLMs.} Now we verify whether nSFT suits stronger VLMs. A similar experiment is done, using a 10k subset of LLaVA-150k. As shown in Table~\ref{tab:ablate-different-models}, stronger models provide higher baseline. In the meanwhile, our nSFT leverages the base model results in most benchmarks for both LLaVA-1.5-13B and LLaVA-NeXT-13B. Our nSFT matches the performance of DPO (SeVa adopted here), both surpassing pure continual SFT in all benchmarks. This has again emphasized the importance of involving \emph{negative supervision}. Together with previous results in LLaVA-1.5-7B in Table~\ref{tab:main-improvement}, we believe that our negative construction pipeline are indeed valid and could probably adapt to more existing VLMs in current publicity. 

\textbf{Compare with RLHF variants.} Besides, we compare nSFT with other RLHF variants, like iterative DPO approach CSR~\cite{csr} and PPO~\cite{LLaVa-RLHF}. To make the comparison fair, we all use a subset of LLaVA-150k, and implement (evaluate) all methods with a same transformer version. As shown in Table~\ref{tab:ablate-nsft-ppo}, iterative DPO can leverage multimodal comprehension \emph{step-by-step}, showing that online DPO samples are beneficial. PPO approach are also effective, but not optimal. In comparison, our nSFT achieves the overall best results, especially on perception benchmarks, showing the power of SFT when negative information are integrated.

\textbf{Components of nSFT.} On top of that, we isolate the component of nSFT: the vision error codebook (VEC) and the chosen responses. Such an experiment is done with a subset of LLaVA-150k, as shown in Table~\ref{tab:ablate-nsft-component}: applying nSFT without a VEC generally leads to the inferior performance, which emphasizes the advantage and necessity of the involving more fine-grained criterions when prompting LLM. Interestingly, we observe that without a chosen information, the model stills achieves similar results. We conjecture that: as the negative construction process \emph{already} regard chosen responses as reference, the final results will becomes less sensitive to its involvement during continual SFT process. 

\textbf{The KL constraint.} At last, we are interested in verifying whether the reference model could help our SFT. We adopt 5k TextCaps as continual learning dataset and add a per-token KL divergency constraints. As shown in the table, the involvement of a KL regularization significantly enhances model's performance, especially for pure continual SFT where model suffers the most alignment tax (e.g., in VQA$^\text{T}$ tasks). With a KL constraint, nSFT further steps upward, showing that the our approach could still be stronger in Table~\ref{tab:main-improvement} if we fully imitate behavior of RLHF.

\section{Conclusion}
In this paper, we advocate the key success of multimodal RLHF lies in its negative supervision, and consequently propose a novel nSFT approach to \emph{fully} matches RLHF. Our hypothesis is strictly verified through theoretical observation and quantitative experiments, as well as fruitful of ablation studies. The nSFT continually aligns VLMs with only one model, making alignment more efficient, too.

As for the limitations, it remains unclear how the proposed nSFT approach suits other LLM area. For example, in natural language processing (NLP) area, the goal of RLHF is to reduce toxic, offensive token output, or to transfer the output generated sentence style, while the goal of multimodal RLHF usually hinges on eliminating the instance- or image-level hallucination. In the future, we might explore whether a similar negative construction style could adapt to RLHF methods in NLP domains.

\appendix

\section{Theoretical derivation}

\subsection{Relations between DPO and SFT.}
In this section, we want to analyze the relations between DPO and SFT, from the \emph{gradient perspective}. We first define the logit of DPO loss function with and without the reference model as $p_{\text{dpo}}$ and $p'_\text{dpo}$, respectively:
\begin{equation}
    p_{\text{dpo}} = \log\frac{\pi_{\theta'}(\boldsymbol{y}_c|\boldsymbol{x})}{\pi_{\text{ref}}(\boldsymbol{y}_c|\boldsymbol{x})}- \log\frac{\pi_{\theta'}(\boldsymbol{y}_r|\boldsymbol{x})}{\pi_{\text{ref}}(\boldsymbol{y}_r|\boldsymbol{x})}\,
\end{equation}
\begin{align}
    p'_{\text{dpo}} &= \log\pi_{\theta'}(\boldsymbol{y}_c|\boldsymbol{x})- \log\pi_{\theta'}(\boldsymbol{y}_r|\boldsymbol{x})\, \\
    &= - (\mathcal{L}_{\text{sft}}(\boldsymbol{y}_c) - \mathcal{L}_{\text{sft}} (\boldsymbol{y}_r))
\end{align}
Then the standard DPO loss function changes to:
\begin{equation}
    \mathcal{L}_{d} = -\log \sigma{(\beta p_\text{dpo})} \,.
\end{equation}
If we take the partial derivation of the DPO loss function to the LLM parameter $\theta'$, we will obtain DPO gradient as:
\begin{align}
    \frac{\partial \mathcal{L}_{d}}{\partial \theta'} &= - \frac{1}{\beta p_\text{dpo}}\frac{\partial (\beta p_\text{dpo})}{\partial \theta'} \\
    &=- \frac{1}{p_\text{dpo}} \frac{\partial p_{\text{dpo}}}{\partial \theta'}\\
    &=- \frac{1}{p_\text{dpo}} \frac{\partial p'_{\text{dpo}}}{\partial \theta'}\\
    &=\frac{1}{p_\text{dpo}}\left[ \frac{\partial \mathcal{L}_{\text{sft}}(\boldsymbol{y}_c)}{\partial \theta'} - \frac{\partial \mathcal{L}_{\text{sft}}(\boldsymbol{y}_r)}{\partial \theta'} \right] \,.
\end{align}

Note that during derivation, $\frac{\partial p_{\text{dpo}}}{\partial \theta'} = \frac{\partial p'_{\text{dpo}}}{\partial \theta'}$ since the refernce model \emph{do not} receive gradient. In the samewhile, the gradient of common SFT loss to the LLM parameter $\theta$ is represented as (we denote the parameter of SFT during continual learning as $\theta'$):
\begin{equation}
    \frac{\partial \mathcal{L}_{\text{sft}}(\boldsymbol{y})}{\partial \theta} \,.
\end{equation}

That is, the DPO gradient is just a linear combination of two SFT gradient (positive response $\boldsymbol{y}_c$ and negative response $\boldsymbol{y}_r$), respectively, with just a dynamic scaling factor $\frac{1}{p'_\text{dpo}}$. This makes their optimization process similar, and can explain the inferior performance brought by the lack of negative supervision in SFT loss.

\subsection{Gradient analysis}
In this subsection, we analyze the gradient direction of DPO loss function towards the chosen response and reject responses, respectively. From Sec. 3, we know that:

\begin{equation}
    t_1 = \frac{\pi_{\theta'}(\boldsymbol{y}_c|\boldsymbol{x})}{\pi_{\text{ref}}(\boldsymbol{y}_c|\boldsymbol{x})}, \quad t_2 = \frac{\pi_{\theta'}(\boldsymbol{y}_r|\boldsymbol{x})}{\pi_{\text{ref}}(\boldsymbol{y}_r|\boldsymbol{x})}\,,
\end{equation}

\begin{equation}
    \left|\frac{\partial \mathcal{L}}{\partial t_1} / \frac{\partial \mathcal{L}}{\partial t_2} \right| = t_2 / t_1 \,.
\end{equation}

During the optimization process, $t_1$ tends to increase and $t_2$ tends to decrease (in order to optimize the final DPO loss), which make the division factor $t_2 / t_1$ less than 1~\cite{limitation-dpo}. As a result, the gradient will be biased towards $t_2$, the negative supervision in DPO loss function. 

Here we want to clarify that this conclusion also trivially holds in our derivation where the reference term is omitted (e.g., the DPO loss function changes to $\mathcal{L}'_d$). In such cases, we can simply let:
\begin{equation}
    t'_1 = \pi_{\theta'}(\boldsymbol{y}_c|\boldsymbol{x}), \quad t'_2 = \pi_{\theta'}(\boldsymbol{y}_r|\boldsymbol{x})\,,
\end{equation}
and derive the same conclusion where the loss function $\mathcal{L}'_d$ will be biased towards optimizing $t'_2$. This will exactly match the basic form of our derivation in Sec. A.1 (how DPO loss is related to SFT loss without a reference term).

\section{Experiment details}

\subsection{How to construct nSFT?}
We now describe the constructed nSFT data in detail. Note that in OCRVQA, the newly constructed conversation length is 2 (manually constructed). In TextCaps and LLaVA-150k, the new constructed conversation length is 5 (GPT-4 is adopted). 

\textbf{OCRVQA.} These dataset are set of book title pages (usually be viewed as object-centric images). As described in our experiment sections. We construct doubled Q-A (question-answer) pairs for each mistake that the model made, and appended these constructed pairs into the tail of the original GT conversation. During our implementation, we found that without the original GT conversation (only the negative constructed pairs are used for training), the training process can be unstable and the performance is unsatisfactory. However, this phenomenon does not apply to TextCaps and LLaVA-150k dataset, where the role of GT information is \emph{not} necessary, as shown in Table 6 in main paper. We conjecture that the constructed conversation in OCRVQA contains few information, as it derives from the wrong answers that only contain one or few tokens that cannot fully describe the whole images. 

\textbf{TextCaps \& LLaVA-150k.} In these two dataset, we adopted GPT-4 to identify the error content in the rejected responses: \emph{the model's original answer without temperature sampling.} The GT annotations are adopted as reference information of the image the guide the identification process. During experiment, we constructed 5 conversations per image, as we found more conversations will increase the overlapping probability with previous constructed data.

\begin{figure}
	\centering
    \includegraphics[width=0.96\linewidth]{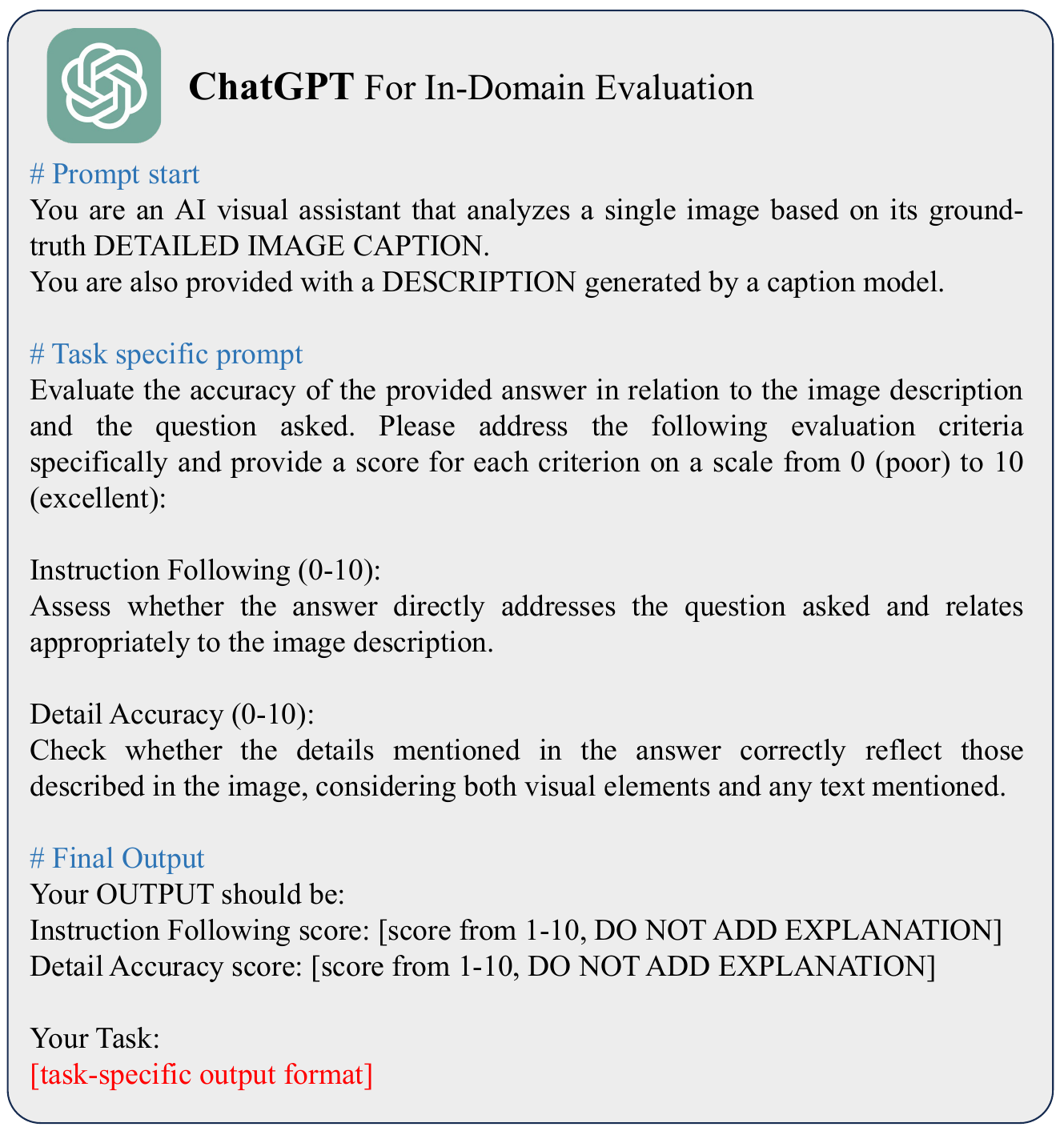}
    \caption{Visualizations of prompt to evaluate in-domain results}
	\label{fig:appendix-prompt-in-domain}
\end{figure}

\subsection{Experimental settings}
Here we will primarily focus on two evaluation metrics.

\textbf{CHAIR.} This benchmark refers to the evaluation metrics adopted in~\cite{chair}. It contains two core metrics:
\begin{equation}
    \text{CHAIR}^i = \frac{|\{\text{hallucinated objects}\}|}{|\{\text{all mentioned objects}\}|}
\end{equation}
\begin{equation}
    \text{CHAIR}^s = \frac{|\{\text{sentence with hallucinated objects}\}|}{|\{\text{all sentence}\}|}
\end{equation}
In experiment, we randomly sample 1,000 COCO validation images, and pair each image with 5 questions of producing detailed captions utilized in LLaVA~\cite{LLM_Llava}. We then use the tools adopted in CHAIR evaluation process to match the object token in MS-COCO~\cite{COCO} and calculate the final results. Finally, we simply average the score obtained by CHAIR$^i$ and CHAIR$^s$:
\begin{equation}
    \text{CHAIR} = (\text{CHAIR}^i + \text{CHAIR}^s) / 2
\end{equation}

\textbf{MMHal.} MMhal evaluation is proposed by~\cite{LLaVa-RLHF}, which aims to evaluate the hallucination ratios from range 0 to 6. Due to the quota limit in the company and heavy evaluation in our experiment, we adopted ChatGPT-3.5 (instead of GPT-4) to evaluate the generated responses. As a result, the number evaluated by ChatGPT-3.5 in Table 1 (in the main paper) and and Table 8 (in the appendix) is slightly higher than the results in GPT-4. To verify whether the version changes will influence the final comparison, we conduct a short experiment, by involving GPT's both version. As shown in Table~\ref{tab:appendix-gpt}, the evaluation score by GPT-3.5/4.0 is different (e.g., the GPT-3.5's is generally higher). However, we have observed a same growing trend with regard to the 4 methods, showing that a replacement of 3.5 version is valid.

\begin{table}
	\small
	\centering
	\begin{tabular}{ccllccccc}
		\toprule[1pt]
		\multirow{2}{*}{GPT version}  &\multicolumn{4}{c}{Alignment methods} \\
         & baseline & GT-DPO  & SeVa & nSFT \\
        \midrule
        ChatGPT3.5 & 2.80 & 2.83 & 2.90 & 3.01 \\
        GPT-4 & 2.04 & 2.05 & 2.12 & 2.20\\
              
        \bottomrule[1pt]
	\end{tabular}
\caption{The evaluation consistency introduced by GPT-3.5/4.0}
\label{tab:appendix-gpt}
\end{table}

\begin{figure*}
	\centering
    \begin{subfigure}{0.43\linewidth}
		\includegraphics[width=0.9\linewidth]{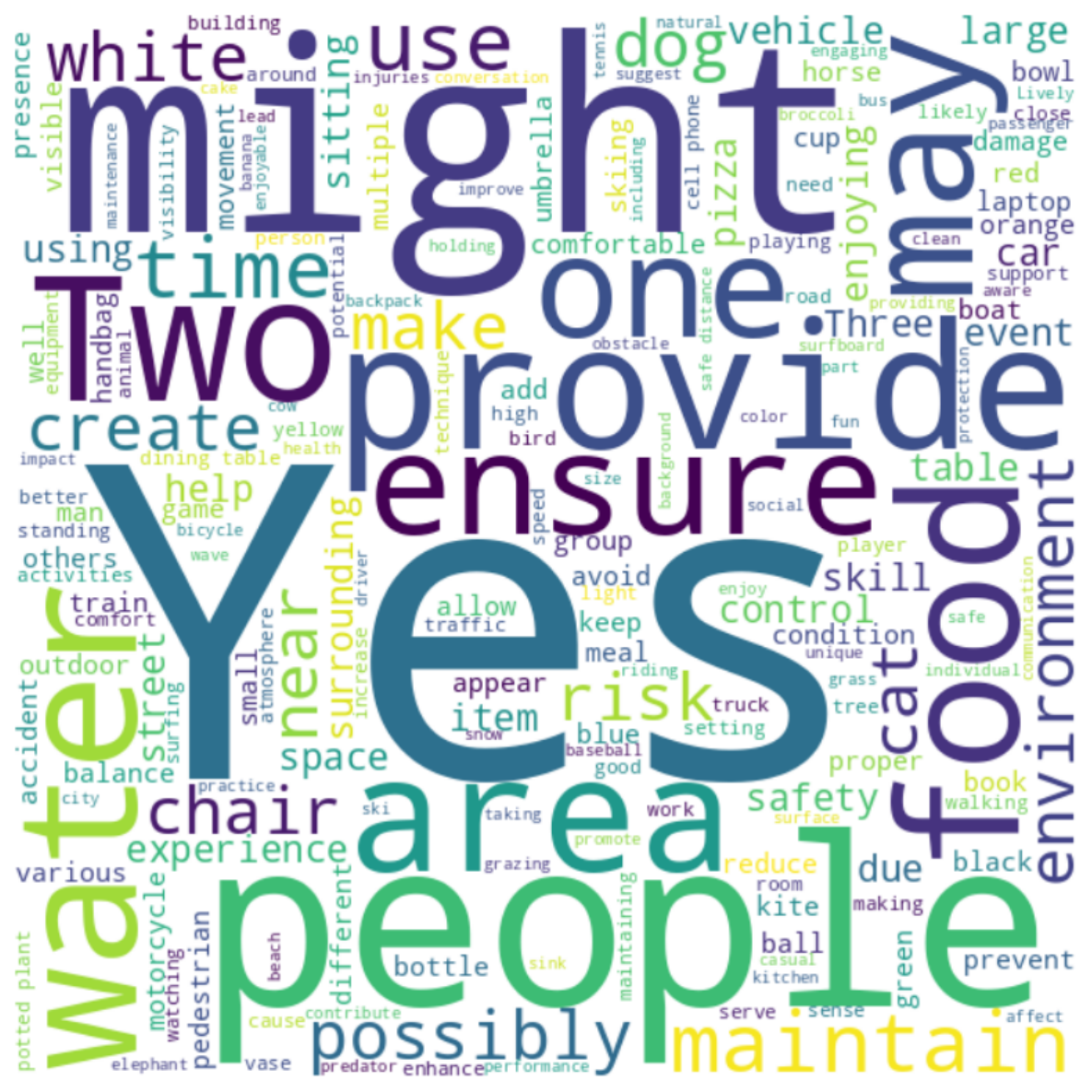}
		\caption{w/o vision error codebook}
    	\label{fig:appendix-wordcloud-degrade}
	\end{subfigure}
	\begin{subfigure}{0.43\linewidth}
		\includegraphics[width=0.9\linewidth]{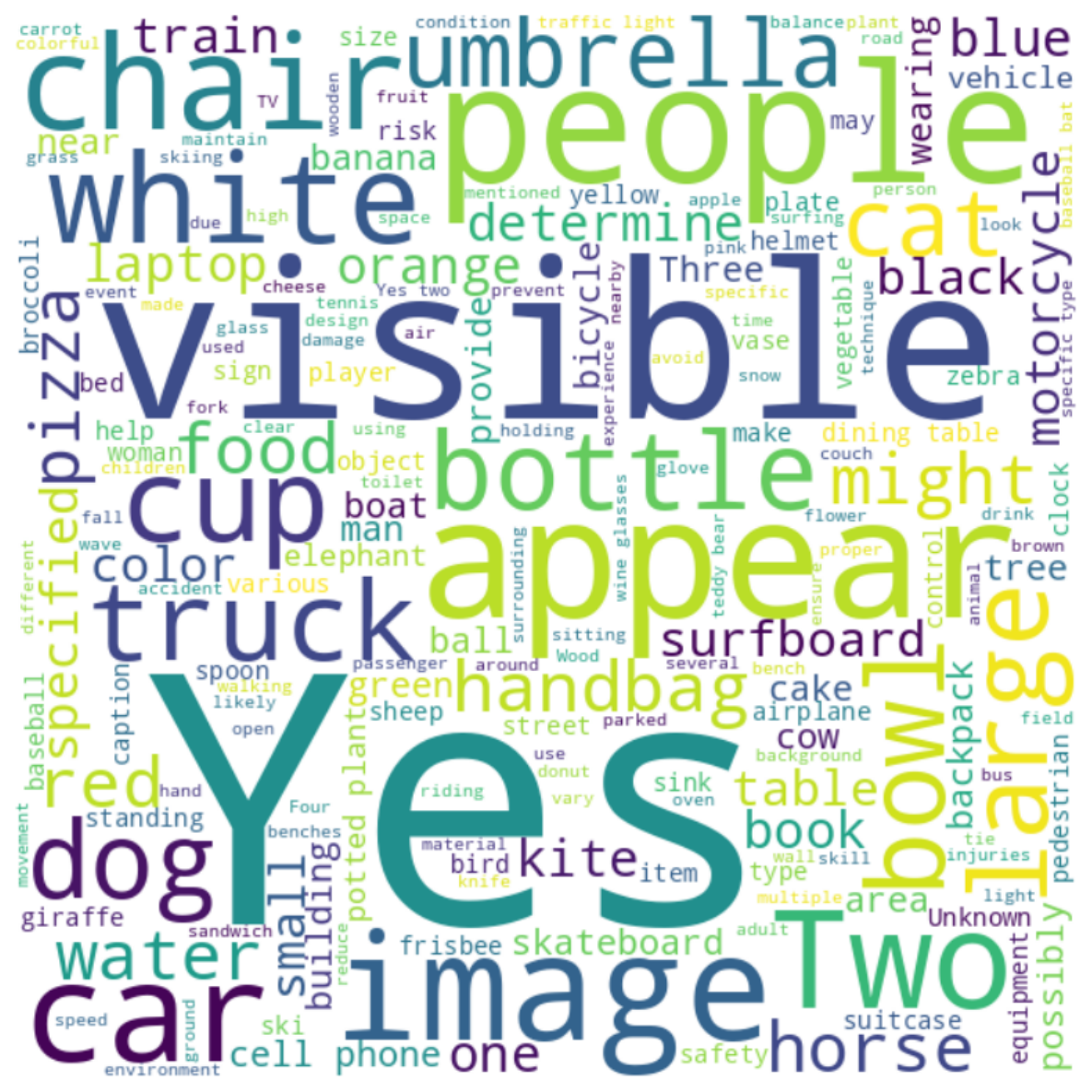}
    	\caption{w/ vision error codebook}
        \label{fig:appendix-wordcloud-nSFT}
    \end{subfigure} 
    \caption{Visualization of the constructed conversation w/o and w/ our vision error codebook, derived from a 5k subset of LLaVA-150k.}
	\label{fig:appendix-wordcloud}
\end{figure*}

\textbf{In-domain evaluation.} The motivation of this experiment in Table 3 is to thoroughly ablate the affect of multimodal RLHF as well as our SFT method. 

Unlike classic computer vision~\cite{QFD}, VLMs benchmarks~\cite{MMBench,MMVet} usually have a large domain shift from its training data~\cite{zhu2024multi,zhu2024llava}, which can induce potential noise that the \emph{core conclusion} might not be emerged. For example, the LLaVA-1.5 continually trained on OCRVQA could probably behave well on books title recognition, but not specialized for benchmarks that relies on image reasoning. In this paper, we first train LLaVA-1.5-7B on 5k subset of OCRVQA, TextCaps and LLaVA-150k, respectively. Then we evaluate each model on its own data source using 500 instances in a \emph{held-out} manner. The evaluation metrics are the instruction following ability and detailed accuracy. As shown in Table 3 in main paper, all methods jointly improve the LLaVA-1.5-7B results, showing that this in-domain dataset design is \emph{indeed} valid. Interestingly, we found the worst accuracy score is relatively higher in RLHF paradigm:  GT-DPO and SeVa obtains the best score in ACC$^w_{10}$, while the metric ACC$_{10}^b$ lean towards our nSFT. We guess this can be attributed to different training paradigm.

In SFT, although we integrate the negative supervision into the final nSFT loss, the overall optimization objective can be \emph{positive}. However, during DPO optimization process, the negative response is deeply integrated into DPO's `logit'. As a result, nSFT implicitly avoid making un-preferred answers, while DPO behave in an explicit way.

\subsection{More data scale comparison}
Please refer to Table~\ref{tab:appendix-comparison} for a whole demonstration. Here we list the results by applying GT-DPO, SeVa~\cite{seva}, SIMA~\cite{sima}, Cont. SFT and nSFT to 3 different databases: OCRVQA~\cite{OCRVQA}, TextCaps and LLaVA-150k. The dataset scale was chosen at 5k and 10k for each specific data source. Although more data scale might further improve the VLM's comprehension ability, we found that this effect becomes weaker when the data scale are beyond 15k-20k. We guess this is due to the data diversity issue. That is, the data for alignment has partially been seen in LLaVA-1.5 SFT stage, which calls for more diverse pretraining or SFT databases. A similar phenomenon can be observed in previous preference alignment articles, where they mostly adopt less than 20k datasource during preference alignment. 

To study how data scale affect alignment is interesting. However, this is beyond the scope of this paper, and we will leave this as future work.

\section{Visualizations}

\subsection{Prompt of in-domain evaluation}
We provide the prompt to evaluate the in-domain results in Fig.~\ref{fig:appendix-prompt-in-domain}. This prompt is sent to GPT-4 to evaluate both the instruction following score, as well as the detailed accuracy score, and the results are shown in Table 3 in main paper.

\subsection{Prompt of nSFT}
\begin{figure*}
	\centering
    \includegraphics[width=0.95\linewidth]{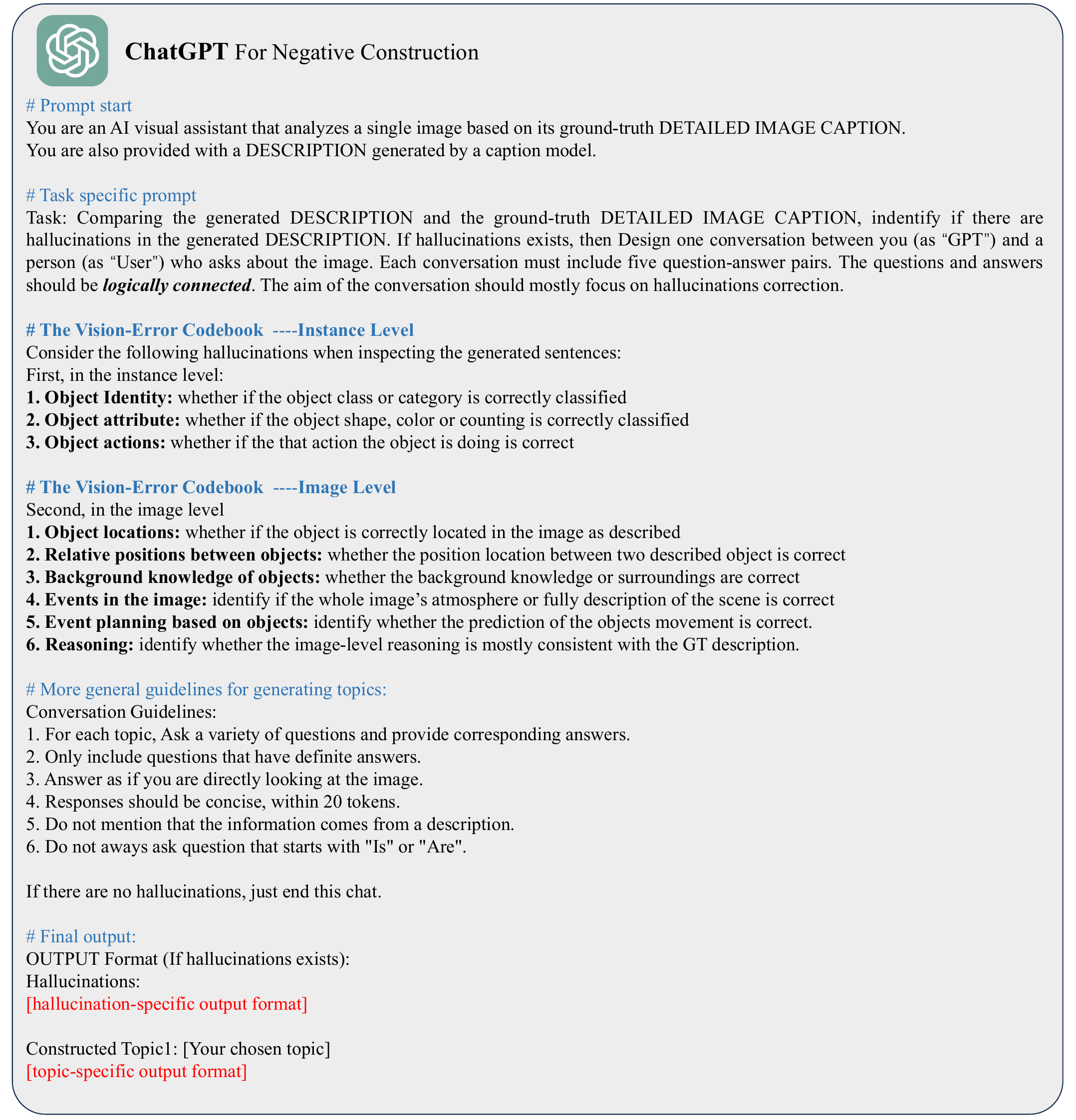}
    \caption{The prompt template used for error identification and conversation reconstruction.}
	\label{fig:appendix-prompt}
\end{figure*}

Please refer to Fig.~\ref{fig:appendix-prompt} for the total prompt to construct the negative supervision. This contains the element of our vision error codebook, which are highlighted with bold phase.

\subsection{Wordcloud visualization}
Please refer to Fig.~\ref{fig:appendix-wordcloud-degrade}-\ref{fig:appendix-wordcloud-nSFT} for the word visualization of our nSFT. The conversation data is sourced from 5k subset of LLaVA-150k. As shown in Fig.~\ref{fig:appendix-wordcloud}, when the conversation is \emph{not} guided by our vision error codebook, it will mostly hinges on non-object phrase, such as `might', `provide'. When we force LLM to focus on specific object type, more object words emerged, such as `truck', `cup', and `chair'.

\subsection{OCRVQA, TextCaps and LLaVA}
\begin{figure*}
	\centering
    \includegraphics[width=0.95\linewidth]{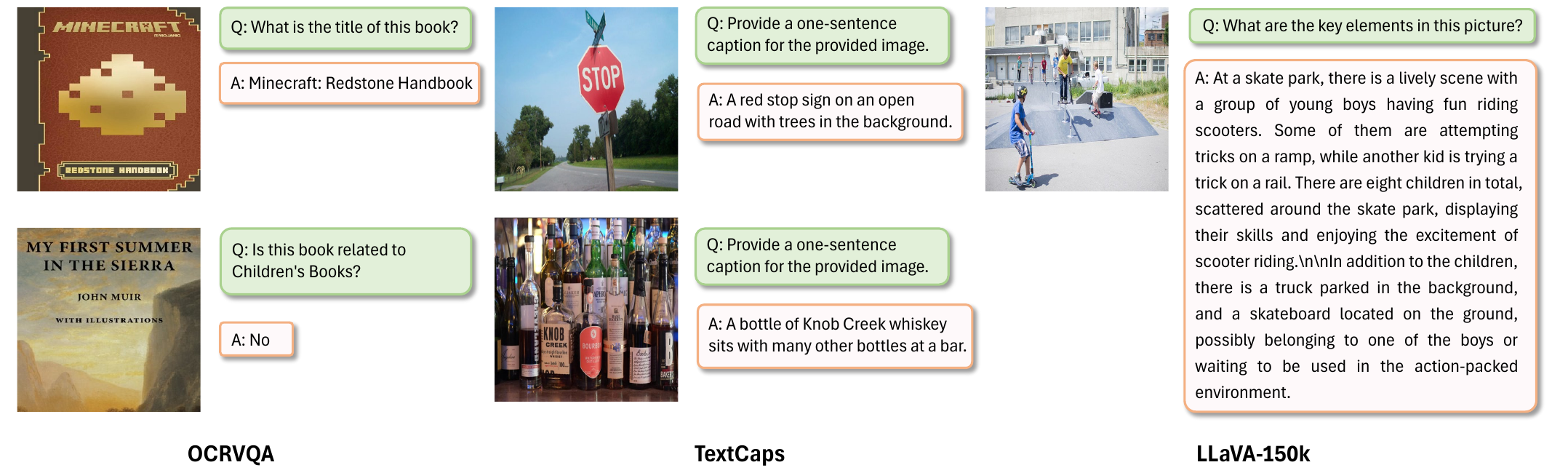}
    \caption{Visualization of groundtruth (GT) captions in OCRVQA, TextCaps and LLaVA-150k dataset. The GT length from left to right has seen a steady growth. For clarity, we omit the suffix of TextCaps question (e.g., `Reference OCR token').}
	\label{fig:appendix-dataset}
\end{figure*}

Please refer to Fig.~\ref{fig:appendix-dataset} to see alignment dataset. Note that the OCRVQA dataset has the shortest response token length, where the responses contains only single word or phrase. The annotations of TextCaps is longer, similar to the length of MS-COCO~\cite{COCO} captions. The captions of LLaVA-150k is much longer, which is constructed by GPT-4 with the original annotations in MS-COCO. 

\subsection{Negative supervision of nSFT}
\begin{figure*}
	\centering
    \includegraphics[width=0.99\linewidth]{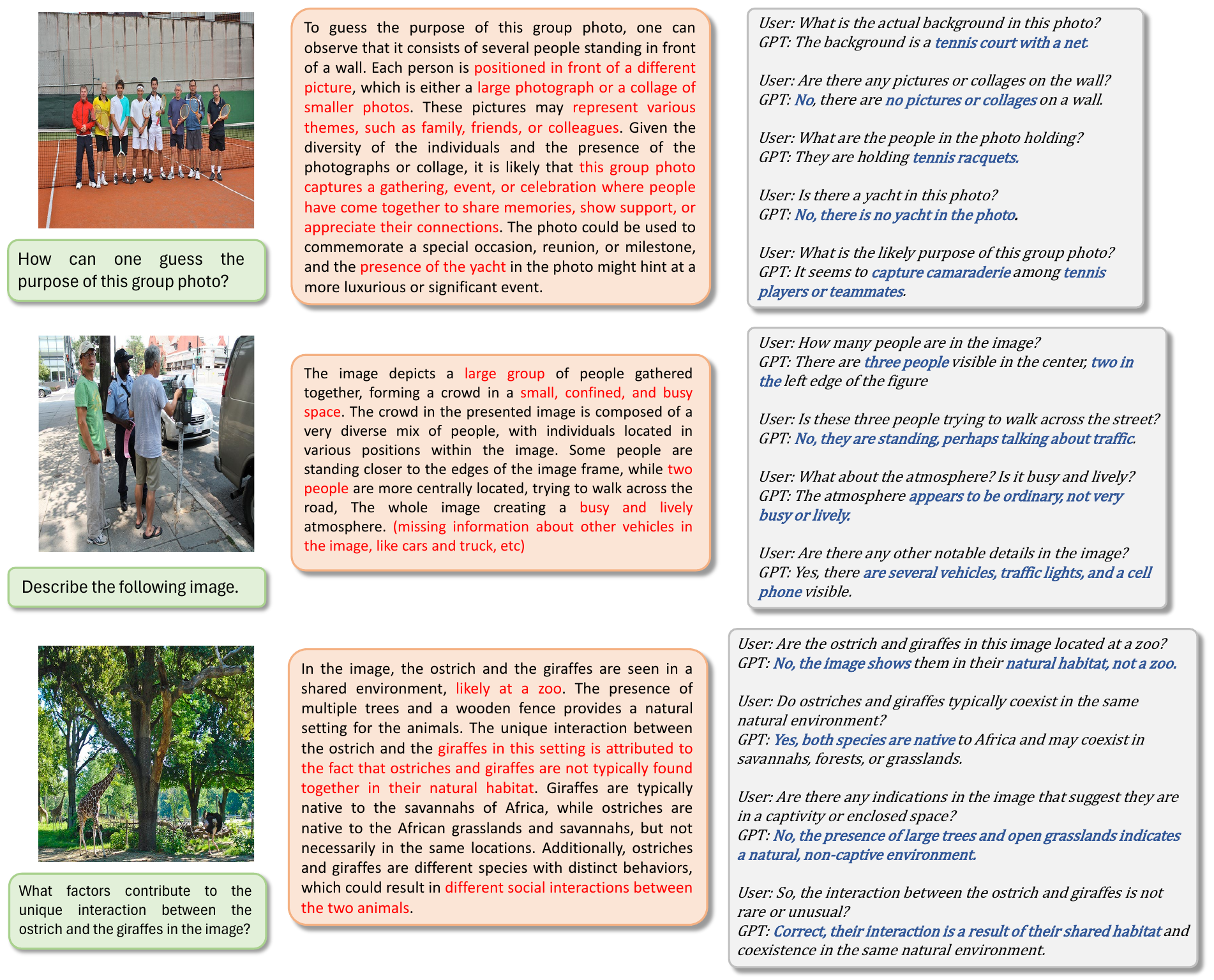}
    \caption{Random sampled cases of our negative constructed supervision. The middle part are the model's original responses to the question, and the right part shows the reconstructed conversations (our nSFT sample). Error content are red color coded. }
	\label{fig:appendix-example}
\end{figure*}

Please refer to Fig.~\ref{fig:appendix-example} for more examples of nSFT. As shown in these figure, models are usually tend to make wrong existence of object (also called image hallucinations). Our nSFT re-inforce these false information by asking the model about the image content. In practical, we balance the `Yes' and `No' ratio in the conversation by randomly erase some `No' answers in the constructed conversations. Empirically this could lead to a more robust result.

\begin{table*}
	\footnotesize
	\centering
	\begin{tabular}{lllllllllllllll}
		\toprule[1pt]
		   \multicolumn{1}{c|}{\multirow{2}{*}{Align. Data}} & \multicolumn{1}{c|}{\multirow{2}{*}{Method}}  & \multicolumn{4}{c|}{Align. tax}& \multicolumn{4}{c|}{Comprehension} & \multicolumn{4}{c}{Hallucinations}\\


      & \multicolumn{1}{|c}{} & \multicolumn{1}{|c}{SQA}  & GQA & VQA$^{\text{T}}$ &  \multicolumn{1}{c|}{total} & MMVet & MME & MMB  &\multicolumn{1}{c|}{total}  & POPE & CHAIR$^{*}$ & MMHal & \multicolumn{1}{c}{total}\\
		\midrule[1pt] 

    \multirow{6}{*}{OCRVQA-5k}       
        & baseline & 66.8 & 62.0 & 58.0 & \gbf{+0.0} & 30.5 & 1510 & 64.3 & \gbf{+0.0} & 85.9 & 32.0 & 2.80 & \gbf{+0.0}  \\
        & GT-DPO & 67.6 & 61.5 & 58.2 & \gbf{+0.5} & 32.3 & 1339 & 63.9 & \gbf{+1.3} & 84.4 & 32.0 & 2.91 & \gbf{+0.3}  \\
        & SeVa & 67.8 & 61.9 & 57.9 & \gbf{+0.8} & 32.2 & 1511 & 64.6 & \gbf{+2.0} & 86.5 & 28.5 & 2.92 & \gbf{+6.1}  \\
        & SIMA & 67.7 & 61.8 & 58.2 & \gbf{+0.9} & 32.8 & 1460 & 64.5 & \gbf{+2.5} & 85.8 & 30.5 & 2.90 & \gbf{+3.1}  \\
        & Cont. SFT & 67.7 & 61.8 & 57.3 & \gbf{+0.0} & 30.8 & 1453 & 63.9 & \rbf{-0.1} & 86.1 & 32.0 & 2.83 & \gbf{+0.7}  \\
        & nSFT & \textbf{68.0} & \textbf{62.0} & \textbf{58.1} & \gbf{+1.3} & \textbf{32.6} & \textbf{1512} & \textbf{64.8} & \gbf{+2.6} & \textbf{86.7} & \textbf{28.2} & \textbf{2.93} & \gbf{+6.8}  \\
        
    \midrule[1pt]

    \multirow{6}{*}{OCRVQA-10k} 
        & baseline & 66.8 & 62.0 & 58.0 & \gbf{+0.0} & 30.5 & 1510 & 64.3 & \gbf{+0.0} & 85.9 & 32.0 & 2.80 & \gbf{+0.0}  \\
        & GT-DPO & 67.8 & 61.4 & 57.7 & \gbf{+0.1} & 32.5 & 1412 & 63.9 & \gbf{+1.6} & 84.3 & 31.5 & 2.90 & \gbf{+0.6}  \\
        & SeVa & 67.6 & 62.0 & 57.5 & \gbf{+0.3} & 32.5 & 1502 & 64.9 & \gbf{+2.6} & 86.6 & 27.3 & 3.00 & \gbf{+8.7}  \\
        & SIMA & 68.0 & 61.9 & 58.2 & \gbf{+1.3} & 32.5 & 1486 & 64.8 & \gbf{+2.5} & 86.2 & 29.4 & 2.93 & \gbf{+5.1}  \\
        & Cont. SFT & 67.9 & 61.7 & 56.9 & \rbf{-0.3} & 33.3 & 1490 & 64.5 & \gbf{+3.0} & 87.0 & 34.0 & 2.76 & \rbf{-1.6}  \\
        & nSFT & \textbf{68.1} & \textbf{62.0} & \textbf{58.1} & \gbf{+1.4} & \textbf{34.0} & \textbf{1515} & \textbf{64.9} & \gbf{+4.1} & \textbf{87.1} & \textbf{26.5} & \textbf{2.93} & \gbf{+8.9}  \\
    \midrule[1pt]
    
    \multirow{6}{*}{TextCaps-5K} 
        & baseline & 66.8 & 62.0 & 58.0 & \gbf{+0.0} & 30.5 & 1510 & 64.3 & \gbf{+0.0} & 85.9 & 32.0 & 2.80 & \gbf{+0.0} \\
        & GT-DPO & 67.7 & 61.9 & 57.8 & \gbf{+0.6} & 33.0 & 1503 & 64.0 & \gbf{+2.2} & 86.3 & 29.6 & 2.83 & \gbf{+3.3} \\
        & SeVa & 67.7 & 61.8 & 57.8 & \gbf{+0.5} & 32.8 & 1498 & 65.0 & \gbf{+3.0} & 86.0 & 27.8 & 2.90 & \gbf{+6.0} \\
        & SIMA & 68.0 & 62.0 & 58.1 & \gbf{+1.3} & 32.8 & 1477 & 65.0 & \gbf{+3.0} & 85.9 & 29.3 & 2.90 & \gbf{+4.4} \\
        & Cont. SFT & 67.3 & 61.5 & 56.6 & \rbf{-1.4} & 32.5 & 1518 & 63.6 & \gbf{+1.3} & 86.3 & 31.5 & 2.91 & \gbf{+2.7} \\
        & nSFT & \textbf{68.1} & \textbf{62.0} & \textbf{58.1} & \gbf{+1.4} & \textbf{33.0} & \textbf{1515} & \textbf{64.8} & \gbf{+3.0} & \textbf{86.8} & \textbf{27.5} & \textbf{2.91} & \gbf{+7.2} \\

    \midrule[1pt]

    \multirow{6}{*}{TextCaps-10K} 
        & baseline & 66.8 & 62.0 & 58.0 & \gbf{+0.0} & 30.5 & 1510 & 64.3 & \gbf{+0.0} & 85.9 & 32.0 & 2.80 & \gbf{+0.0} \\
        & GT-DPO & 68.0 & 61.7 & 57.5 & \gbf{+0.4} & 34.2 & 1500 & 64.2 & \gbf{+3.6} & 86.5 & 29.2 & 2.83 & \gbf{+3.9} \\
        & SeVa & 68.1 & 61.7 & 57.8 & \gbf{+0.8} & 34.6 & 1480 & 65.0 & \gbf{+4.8} & 86.3 & 26.3 & 2.90 & \gbf{+7.8} \\
        & SIMA & 68.0 & 62.1 & 58.0 & \gbf{+1.3} & 32.2 & 1473 & 64.9 & \gbf{+2.3} & 85.9 & 27.6 & 2.87 & \gbf{+5.6} \\
        & Cont. SFT & 66.9 & 61.3 & 56.6 & \rbf{-2.0} & 31.0 & 1520 & 64.4 & \gbf{+0.6} & 86.3 & 30.5 & 2.83 & \gbf{+2.4} \\
        & nSFT & \textbf{68.4} & \textbf{62.3} & \textbf{58.2} & \gbf{+2.1} & \textbf{33.7} & \textbf{1521} & \textbf{65.3} & \gbf{+4.2} & \textbf{87.2} & \textbf{26.2} & \textbf{2.97} & \gbf{+9.9} \\

    \midrule[1pt]

    \multirow{6}{*}{LLaVAData-5k}  
        & baseline & 66.8 & 62.0 & 58.0 & \gbf{+0.0} & 30.5 & 1510 & 64.3 & \gbf{+0.0} & 85.9 & 32.0 & 2.80 & \gbf{+0.0} \\
        & GT-DPO & 67.9 & 61.9 & 58.0 & \gbf{+1.0} & 32.6 & 1512 & 64.1 & \gbf{+1.9} & 86.1 & 28.5 & 2.94 & \gbf{+6.0} \\
        & SeVa & 67.6 & 61.7 & 58.2 & \gbf{+0.7} & 32.6 & 1505 & 64.7 & \gbf{+2.5} & 86.0 & 28.3 & 2.93 & \gbf{+6.0} \\
        & SIMA & 67.9 & \textbf{62.1} & 58.2 & \gbf{+1.4} & 32.1 & 1505 & 64.5 & \gbf{+1.8} & 86.3 & 28.1 & 2.95 & \gbf{+6.8} \\
        & Cont. SFT & 67.5 & 61.0 & 56.7 & \rbf{-1.6} & 31.0 & 1475 & 63.5 & \rbf{-0.3} & 85.6 & 29.5 & 2.82 & \gbf{+2.5} \\
        & nSFT & \textbf{68.2} & 61.9 & \textbf{58.2} & \gbf{+1.5} & \textbf{33.0} & \textbf{1533} & \textbf{65.0} & \gbf{+3.2} & \textbf{86.5} & \textbf{27.2} & \textbf{2.99} & \gbf{+8.6} \\
    \midrule[1pt]

    \multirow{6}{*}{LLaVAData-10k} 
        & baseline & 66.8 & 62.0 & 58.0 & \gbf{+0.0} & 30.5 & 1510 & 64.3 & \gbf{+0.0} & 85.9 & 32.0 & 2.80 & \gbf{+0.0} \\
        & GT-DPO & 68.1 & 61.6 & 57.6 & \gbf{+0.5} & 33.9 & 1497 & 63.9 & \gbf{+3.0} & 85.9 & 30.7 & 2.80 & \gbf{+1.3} \\
        & SeVa & 67.5 & 61.4 & 58.0 & \gbf{+0.1} & 32.5 & 1490 & 64.7 & \gbf{+2.4} & 85.6 & 28.2 & 2.94 & \gbf{+5.8} \\
        & SIMA & 67.9 & 62.2 & 58.2 & \gbf{+1.5} & 32.1 & 1511 & 64.9 & \gbf{+2.2} & 86.9 & 26.2 & 2.97 & \gbf{+9.6} \\
        & Cont. SFT & 67.1 & 60.9 & 57.0 & \rbf{-1.8} & 31.2 & 1480 & 64.0 & \gbf{+0.4} & 86.3 & 29.1 & 2.91 & \gbf{+5.1} \\
        & nSFT & \textbf{68.4} & \textbf{62.3} & \textbf{58.4} & \gbf{+2.3} & \textbf{34.2} & \textbf{1550} & \textbf{65.2} & \gbf{+4.6} & \textbf{87.4} & \textbf{25.4} & \textbf{3.02} & \gbf{+11.8} \\
  
		\bottomrule[1pt]
	\end{tabular}
\caption{Nine benchmark results by applying 5 continual learning methods. We list the outcomes using alignment data of 5k and 10k.}
\label{tab:appendix-comparison}
\end{table*}

\clearpage

\clearpage
\clearpage



{
    \small
    \bibliographystyle{ieeenat_fullname}
    \bibliography{main}

\begin{thebibliography}{48}
\providecommand{\natexlab}[1]{#1}
\providecommand{\url}[1]{\texttt{#1}}
\expandafter\ifx\csname urlstyle\endcsname\relax
  \providecommand{\doi}[1]{doi: #1}\else
  \providecommand{\doi}{doi: \begingroup \urlstyle{rm}\Url}\fi

\bibitem[Bai et~al.(2023)Bai, Bai, Yang, Wang, Tan, Wang, Lin, Zhou, and Zhou]{Qwen}
Jinze Bai, Shuai Bai, Shusheng Yang, Shijie Wang, Sinan Tan, Peng Wang, Junyang Lin, Chang Zhou, and Jingren Zhou.
\newblock Qwen-vl: A versatile vision-language model for understanding, localization, text reading, and beyond.
\newblock 2023.

\bibitem[Chen et~al.(2023)Chen, Li, Dong, Zhang, He, Wang, Zhao, and Lin]{ShareGPT4V}
Lin Chen, Jisong Li, Xiaoyi Dong, Pan Zhang, Conghui He, Jiaqi Wang, Feng Zhao, and Dahua Lin.
\newblock Sharegpt4v: Improving large multi-modal models with better captions.
\newblock \emph{arXiv preprint arXiv:2311.12793}, 2023.

\bibitem[Chen et~al.()Chen, Deng, Yuan, Ji, and Gu]{self-play}
Zixiang Chen, Yihe Deng, Huizhuo Yuan, Kaixuan Ji, and Quanquan Gu.
\newblock Self-play fine-tuning converts weak language models to strong language models.
\newblock In \emph{Forty-first International Conference on Machine Learning}.

\bibitem[Chen et~al.(2024)Chen, Wang, Tian, Ye, Gao, Cui, Tong, Hu, Luo, Ma, et~al.]{intervl1.5}
Zhe Chen, Weiyun Wang, Hao Tian, Shenglong Ye, Zhangwei Gao, Erfei Cui, Wenwen Tong, Kongzhi Hu, Jiapeng Luo, Zheng Ma, et~al.
\newblock How far are we to gpt-4v? closing the gap to commercial multimodal models with open-source suites.
\newblock \emph{arXiv preprint arXiv:2404.16821}, 2024.

\bibitem[Chung et~al.(2024)Chung, Hou, Longpre, Zoph, Tay, Fedus, Li, Wang, Dehghani, Brahma, et~al.]{sft-nlp-scale-instruct}
Hyung~Won Chung, Le Hou, Shayne Longpre, Barret Zoph, Yi Tay, William Fedus, Yunxuan Li, Xuezhi Wang, Mostafa Dehghani, Siddhartha Brahma, et~al.
\newblock Scaling instruction-finetuned language models.
\newblock \emph{Journal of Machine Learning Research}, 25\penalty0 (70):\penalty0 1--53, 2024.

\bibitem[Dong et~al.(2024)Dong, Han, Peng, Qi, Ge, Yang, Zhao, Sun, Zhou, Wei, Kong, Zhang, Ma, and Yi]{DreamLLM}
Runpei Dong, Chunrui Han, Yuang Peng, Zekun Qi, Zheng Ge, Jinrong Yang, Liang Zhao, Jianjian Sun, Hongyu Zhou, Haoran Wei, Xiangwen Kong, Xiangyu Zhang, Kaisheng Ma, and Li Yi.
\newblock Dream{LLM}: Synergistic multimodal comprehension and creation.
\newblock In \emph{The Twelfth International Conference on Learning Representations}, 2024.

\bibitem[Feng et~al.(2024)Feng, Qin, Huang, Zhang, and Lei]{limitation-dpo}
Duanyu Feng, Bowen Qin, Chen Huang, Zheng Zhang, and Wenqiang Lei.
\newblock Towards analyzing and understanding the limitations of dpo: A theoretical perspective.
\newblock \emph{arXiv preprint arXiv:2404.04626}, 2024.

\bibitem[Hudson and Manning(2019)]{gqa}
Drew~A Hudson and Christopher~D Manning.
\newblock {GQA}: A new dataset for real-world visual reasoning and compositional question answering.
\newblock In \emph{Proceedings of the IEEE/CVF conference on computer vision and pattern recognition}, pages 6700--6709, 2019.

\bibitem[Li et~al.(2023{\natexlab{a}})Li, Li, Savarese, and Hoi]{LLM_BLIP2}
Junnan Li, Dongxu Li, Silvio Savarese, and Steven Hoi.
\newblock Blip-2: Bootstrapping language-image pre-training with frozen image encoders and large language models.
\newblock In \emph{International conference on machine learning}, pages 19730--19742. PMLR, 2023{\natexlab{a}}.

\bibitem[Li et~al.(2024{\natexlab{a}})Li, Lin, and Pei]{VLLM_DPO}
Shengzhi Li, Rongyu Lin, and Shichao Pei.
\newblock Multi-modal preference alignment remedies regression of visual instruction tuning on language model.
\newblock \emph{arXiv preprint arXiv:2402.10884}, 2024{\natexlab{a}}.

\bibitem[Li et~al.(2024{\natexlab{b}})Li, Lin, and Pei]{vDPO_Gemini}
Shengzhi Li, Rongyu Lin, and Shichao Pei.
\newblock Multi-modal preference alignment remedies regression of visual instruction tuning on language model.
\newblock \emph{arXiv preprint arXiv:2402.10884}, 2024{\natexlab{b}}.

\bibitem[Li et~al.(2023{\natexlab{b}})Li, Du, Zhou, Wang, Zhao, and Wen]{POPE}
Yifan Li, Yifan Du, Kun Zhou, Jinpeng Wang, Wayne~Xin Zhao, and Ji-Rong Wen.
\newblock Evaluating object hallucination in large vision-language models.
\newblock In \emph{Proceedings of the 2023 Conference on Empirical Methods in Natural Language Processing}, pages 292--305, 2023{\natexlab{b}}.

\bibitem[Lin et~al.(2014)Lin, Maire, Belongie, Hays, Perona, Ramanan, Doll{\'a}r, and Zitnick]{COCO}
Tsung-Yi Lin, Michael Maire, Serge Belongie, James Hays, Pietro Perona, Deva Ramanan, Piotr Doll{\'a}r, and C~Lawrence Zitnick.
\newblock Microsoft coco: Common objects in context.
\newblock In \emph{Computer Vision--ECCV 2014: 13th European Conference, Zurich, Switzerland, September 6-12, 2014, Proceedings, Part V 13}, pages 740--755. Springer, 2014.

\bibitem[Liu et~al.(2023{\natexlab{a}})Liu, Li, Li, and Lee]{LLaVa1.5}
Haotian Liu, Chunyuan Li, Yuheng Li, and Yong~Jae Lee.
\newblock Improved baselines with visual instruction tuning.
\newblock \emph{arXiv preprint arXiv:2310.03744}, 2023{\natexlab{a}}.

\bibitem[Liu et~al.(2024{\natexlab{a}})Liu, Li, Li, Li, Zhang, Shen, and Lee]{llava-next}
Haotian Liu, Chunyuan Li, Yuheng Li, Bo Li, Yuanhan Zhang, Sheng Shen, and Yong~Jae Lee.
\newblock Llava-next: Improved reasoning, ocr, and world knowledge, 2024{\natexlab{a}}.

\bibitem[Liu et~al.(2024{\natexlab{b}})Liu, Li, Wu, and Lee]{LLM_Llava}
Haotian Liu, Chunyuan Li, Qingyang Wu, and Yong~Jae Lee.
\newblock Visual instruction tuning.
\newblock \emph{Advances in neural information processing systems}, 36, 2024{\natexlab{b}}.

\bibitem[Liu et~al.(2023{\natexlab{b}})Liu, Duan, Zhang, Li, Zhang, Zhao, Yuan, Wang, He, Liu, et~al.]{MMBench}
Yuan Liu, Haodong Duan, Yuanhan Zhang, Bo Li, Songyang Zhang, Wangbo Zhao, Yike Yuan, Jiaqi Wang, Conghui He, Ziwei Liu, et~al.
\newblock Mmbench: Is your multi-modal model an all-around player?
\newblock \emph{arXiv preprint arXiv:2307.06281}, 2023{\natexlab{b}}.

\bibitem[Liu et~al.(2024{\natexlab{c}})Liu, Lu, Zhang, Liu, Guo, Yang, Blanchet, and Wang]{sft-implicit-regular-dpo}
Zhihan Liu, Miao Lu, Shenao Zhang, Boyi Liu, Hongyi Guo, Yingxiang Yang, Jose Blanchet, and Zhaoran Wang.
\newblock Provably mitigating overoptimization in rlhf: Your sft loss is implicitly an adversarial regularizer.
\newblock \emph{arXiv preprint arXiv:2405.16436}, 2024{\natexlab{c}}.

\bibitem[Lu et~al.(2024)Lu, Yu, Huang, Fan, Lin, and Zhou]{merge-optimizer-dpo}
Keming Lu, Bowen Yu, Fei Huang, Yang Fan, Runji Lin, and Chang Zhou.
\newblock Online merging optimizers for boosting rewards and mitigating tax in alignment.
\newblock \emph{arXiv preprint arXiv:2405.17931}, 2024.

\bibitem[Lu et~al.(2022)Lu, Mishra, Xia, Qiu, Chang, Zhu, Tafjord, Clark, and Kalyan]{sqa}
Pan Lu, Swaroop Mishra, Tony Xia, Liang Qiu, Kai-Wei Chang, Song-Chun Zhu, Oyvind Tafjord, Peter Clark, and Ashwin Kalyan.
\newblock Learn to explain: Multimodal reasoning via thought chains for science question answering.
\newblock In \emph{The 36th Conference on Neural Information Processing Systems (NeurIPS)}, 2022.

\bibitem[Mishra et~al.(2019)Mishra, Shekhar, Singh, and Chakraborty]{OCRVQA}
Anand Mishra, Shashank Shekhar, Ajeet~Kumar Singh, and Anirban Chakraborty.
\newblock Ocr-vqa: Visual question answering by reading text in images.
\newblock In \emph{2019 international conference on document analysis and recognition (ICDAR)}, pages 947--952. IEEE, 2019.

\bibitem[Pi et~al.(2024)Pi, Han, Xiong, Zhang, Liu, Pan, and Zhang]{bpo}
Renjie Pi, Tianyang Han, Wei Xiong, Jipeng Zhang, Runtao Liu, Rui Pan, and Tong Zhang.
\newblock Strengthening multimodal large language model with bootstrapped preference optimization.
\newblock \emph{arXiv preprint arXiv:2403.08730}, 2024.

\bibitem[Rafailov et~al.(2024)Rafailov, Sharma, Mitchell, Manning, Ermon, and Finn]{DPO}
Rafael Rafailov, Archit Sharma, Eric Mitchell, Christopher~D Manning, Stefano Ermon, and Chelsea Finn.
\newblock Direct preference optimization: Your language model is secretly a reward model.
\newblock 2024.

\bibitem[Rohrbach et~al.(2018)Rohrbach, Hendricks, Burns, Darrell, and Saenko]{chair}
Anna Rohrbach, Lisa~Anne Hendricks, Kaylee Burns, Trevor Darrell, and Kate Saenko.
\newblock Object hallucination in image captioning.
\newblock In \emph{Proceedings of the 2018 Conference on Empirical Methods in Natural Language Processing}, pages 4035--4045, 2018.

\bibitem[Sanh et~al.()Sanh, Webson, Raffel, Bach, Sutawika, Alyafeai, Chaffin, Stiegler, Raja, Dey, et~al.]{sft-nlp-multitask}
Victor Sanh, Albert Webson, Colin Raffel, Stephen Bach, Lintang Sutawika, Zaid Alyafeai, Antoine Chaffin, Arnaud Stiegler, Arun Raja, Manan Dey, et~al.
\newblock Multitask prompted training enables zero-shot task generalization.
\newblock In \emph{International Conference on Learning Representations}.

\bibitem[Schulman et~al.(2017)Schulman, Wolski, Dhariwal, Radford, and Klimov]{PPO}
John Schulman, Filip Wolski, Prafulla Dhariwal, Alec Radford, and Oleg Klimov.
\newblock Proximal policy optimization algorithms.
\newblock \emph{arXiv preprint arXiv:1707.06347}, 2017.

\bibitem[Singh et~al.(2019)Singh, Natarajan, Shah, Jiang, Chen, Batra, Parikh, and Rohrbach]{TextVQA}
Amanpreet Singh, Vivek Natarajan, Meet Shah, Yu Jiang, Xinlei Chen, Dhruv Batra, Devi Parikh, and Marcus Rohrbach.
\newblock Towards vqa models that can read.
\newblock In \emph{Proceedings of the IEEE/CVF conference on computer vision and pattern recognition}, pages 8317--8326, 2019.

\bibitem[Sun et~al.(2024)Sun, Zhang, Chen, Zhang, Sang, Zhang, Wang, and Li]{arcana}
Yanpeng Sun, Huaxin Zhang, Qiang Chen, Xinyu Zhang, Nong Sang, Gang Zhang, Jingdong Wang, and Zechao Li.
\newblock Improving multi-modal large language model through boosting vision capabilities.
\newblock \emph{arXiv preprint arXiv:2410.13733}, 2024.

\bibitem[Sun et~al.(2023)Sun, Shen, Cao, Liu, Li, Shen, Gan, Gui, Wang, Yang, et~al.]{LLaVa-RLHF}
Zhiqing Sun, Sheng Shen, Shengcao Cao, Haotian Liu, Chunyuan Li, Yikang Shen, Chuang Gan, Liang-Yan Gui, Yu-Xiong Wang, Yiming Yang, et~al.
\newblock Aligning large multimodal models with factually augmented rlhf.
\newblock \emph{arXiv preprint arXiv:2309.14525}, 2023.

\bibitem[Wang et~al.(2023)Wang, Lv, Yu, Hong, Qi, Wang, Ji, Yang, Zhao, Song, et~al.]{cogvlm}
Weihan Wang, Qingsong Lv, Wenmeng Yu, Wenyi Hong, Ji Qi, Yan Wang, Junhui Ji, Zhuoyi Yang, Lei Zhao, Xixuan Song, et~al.
\newblock Cogvlm: Visual expert for pretrained language models.
\newblock \emph{arXiv preprint arXiv:2311.03079}, 2023.

\bibitem[Wang et~al.(2024)Wang, Chen, Wang, Zhou, Zhou, Yao, Zhou, Goldstein, Bhatia, Huang, et~al.]{sima}
Xiyao Wang, Jiuhai Chen, Zhaoyang Wang, Yuhang Zhou, Yiyang Zhou, Huaxiu Yao, Tianyi Zhou, Tom Goldstein, Parminder Bhatia, Furong Huang, et~al.
\newblock Enhancing visual-language modality alignment in large vision language models via self-improvement.
\newblock \emph{arXiv preprint arXiv:2405.15973}, 2024.

\bibitem[Wei et~al.(2023)Wei, Kong, Chen, Zhao, Ge, Yang, Sun, Han, and Zhang]{vary}
Haoran Wei, Lingyu Kong, Jinyue Chen, Liang Zhao, Zheng Ge, Jinrong Yang, Jianjian Sun, Chunrui Han, and Xiangyu Zhang.
\newblock Vary: Scaling up the vision vocabulary for large vision-language models.
\newblock \emph{arXiv preprint arXiv:2312.06109}, 2023.

\bibitem[Wei et~al.()Wei, Bosma, Zhao, Guu, Yu, Lester, Du, Dai, and Le]{sft-nlp-finetuned-llm}
Jason Wei, Maarten Bosma, Vincent Zhao, Kelvin Guu, Adams~Wei Yu, Brian Lester, Nan Du, Andrew~M Dai, and Quoc~V Le.
\newblock Finetuned language models are zero-shot learners.
\newblock In \emph{International Conference on Learning Representations}.

\bibitem[Wolf et~al.(2020)Wolf, Debut, Sanh, Chaumond, Delangue, Moi, Cistac, Rault, Louf, Funtowicz, Davison, Shleifer, von Platen, Ma, Jernite, Plu, Xu, Scao, Gugger, Drame, Lhoest, and Rush]{transformers}
Thomas Wolf, Lysandre Debut, Victor Sanh, Julien Chaumond, Clement Delangue, Anthony Moi, Pierric Cistac, Tim Rault, Rémi Louf, Morgan Funtowicz, Joe Davison, Sam Shleifer, Patrick von Platen, Clara Ma, Yacine Jernite, Julien Plu, Canwen Xu, Teven~Le Scao, Sylvain Gugger, Mariama Drame, Quentin Lhoest, and Alexander~M. Rush.
\newblock Transformers: State-of-the-art natural language processing.
\newblock In \emph{Proceedings of the 2020 Conference on Empirical Methods in Natural Language Processing: System Demonstrations}, pages 38--45, Online, 2020. Association for Computational Linguistics.

\bibitem[Wu et~al.(2023)Wu, Yin, Qi, Wang, Tang, and Duan]{ChatGPT}
Chenfei Wu, Shengming Yin, Weizhen Qi, Xiaodong Wang, Zecheng Tang, and Nan Duan.
\newblock Visual chatgpt: Talking, drawing and editing with visual foundation models.
\newblock \emph{arXiv preprint arXiv:2303.04671}, 2023.

\bibitem[Xu et~al.()Xu, Fu, Gao, Ye, Liu, Mei, Wang, Yu, and Wu]{dpo-or-ppo}
Shusheng Xu, Wei Fu, Jiaxuan Gao, Wenjie Ye, Weilin Liu, Zhiyu Mei, Guangju Wang, Chao Yu, and Yi Wu.
\newblock Is dpo superior to ppo for llm alignment? a comprehensive study.
\newblock In \emph{Forty-first International Conference on Machine Learning}.

\bibitem[Yu et~al.(2023)Yu, Yang, Li, Wang, Lin, Liu, Wang, and Wang]{MMVet}
Weihao Yu, Zhengyuan Yang, Linjie Li, Jianfeng Wang, Kevin Lin, Zicheng Liu, Xinchao Wang, and Lijuan Wang.
\newblock Mm-vet: Evaluating large multimodal models for integrated capabilities.
\newblock \emph{arXiv preprint arXiv:2308.02490}, 2023.

\bibitem[Zeng et~al.()Zeng, Liu, Ma, Yang, Zhang, and Wang]{token-dpo}
Yongcheng Zeng, Guoqing Liu, Weiyu Ma, Ning Yang, Haifeng Zhang, and Jun Wang.
\newblock Token-level direct preference optimization.
\newblock In \emph{Forty-first International Conference on Machine Learning}.

\bibitem[Zhao et~al.(2023{\natexlab{a}})Zhao, Yu, Ge, Yang, Wei, Zhou, Sun, Peng, Dong, Han, et~al.]{chatspot}
Liang Zhao, En Yu, Zheng Ge, Jinrong Yang, Haoran Wei, Hongyu Zhou, Jianjian Sun, Yuang Peng, Runpei Dong, Chunrui Han, et~al.
\newblock Chatspot: Bootstrapping multimodal llms via precise referring instruction tuning.
\newblock \emph{arXiv preprint arXiv:2307.09474}, 2023{\natexlab{a}}.

\bibitem[Zhao et~al.(2023{\natexlab{b}})Zhao, Wang, Ouyang, Dong, Wang, and He]{HA-DPO}
Zhiyuan Zhao, Bin Wang, Linke Ouyang, Xiaoyi Dong, Jiaqi Wang, and Conghui He.
\newblock Beyond hallucinations: Enhancing lvlms through hallucination-aware direct preference optimization.
\newblock \emph{arXiv preprint arXiv:2311.16839}, 2023{\natexlab{b}}.

\bibitem[Zhou et~al.(2024{\natexlab{a}})Zhou, Liu, Xu, Iyer, Sun, Mao, Ma, Efrat, Yu, Yu, et~al.]{sft-nlp-less-more}
Chunting Zhou, Pengfei Liu, Puxin Xu, Srinivasan Iyer, Jiao Sun, Yuning Mao, Xuezhe Ma, Avia Efrat, Ping Yu, Lili Yu, et~al.
\newblock Lima: Less is more for alignment.
\newblock \emph{Advances in Neural Information Processing Systems}, 36, 2024{\natexlab{a}}.

\bibitem[Zhou et~al.(2024{\natexlab{b}})Zhou, Fan, Cheng, Yang, Chen, Cui, Wang, Li, Zhang, and Yao]{csr}
Yiyang Zhou, Zhiyuan Fan, Dongjie Cheng, Sihan Yang, Zhaorun Chen, Chenhang Cui, Xiyao Wang, Yun Li, Linjun Zhang, and Huaxiu Yao.
\newblock Calibrated self-rewarding vision language models.
\newblock \emph{arXiv preprint arXiv:2405.14622}, 2024{\natexlab{b}}.

\bibitem[Zhu et~al.(2023{\natexlab{a}})Zhu, Chen, Shen, Li, and Elhoseiny]{LLM_MiniGPT4}
Deyao Zhu, Jun Chen, Xiaoqian Shen, Xiang Li, and Mohamed Elhoseiny.
\newblock Minigpt-4: Enhancing vision-language understanding with advanced large language models.
\newblock \emph{arXiv preprint arXiv:2304.10592}, 2023{\natexlab{a}}.

\bibitem[Zhu et~al.(2023{\natexlab{b}})Zhu, Fu, and Wu]{mls}
Ke Zhu, Minghao Fu, and Jianxin Wu.
\newblock Multi-label self-supervised learning with scene images.
\newblock In \emph{Proceedings of the IEEE/CVF International Conference on Computer Vision}, pages 6694--6703, 2023{\natexlab{b}}.

\bibitem[Zhu et~al.(2023{\natexlab{c}})Zhu, He, and Wu]{QFD}
Ke Zhu, Yin-Yin He, and Jianxin Wu.
\newblock Quantized feature distillation for network quantization.
\newblock In \emph{Proceedings of the AAAI Conference on Artificial Intelligence}, pages 11452--11460, 2023{\natexlab{c}}.

\bibitem[Zhu et~al.(2024{\natexlab{a}})Zhu, Zhao, Ge, and Zhang]{seva}
Ke Zhu, Liang Zhao, Zheng Ge, and Xiangyu Zhang.
\newblock Self-supervised visual preference alignment.
\newblock \emph{arXiv preprint arXiv:2404.10501}, 2024{\natexlab{a}}.

\bibitem[Zhu et~al.(2024{\natexlab{b}})Zhu, Zhu, Liu, Xu, and Peng]{zhu2024llava}
Yichen Zhu, Minjie Zhu, Ning Liu, Zhiyuan Xu, and Yaxin Peng.
\newblock Llava-phi: Efficient multi-modal assistant with small language model.
\newblock In \emph{Proceedings of the 1st International Workshop on Efficient Multimedia Computing under Limited}, pages 18--22, 2024{\natexlab{b}}.

\bibitem[Zhu et~al.(2024{\natexlab{c}})Zhu, Xu, Chen, Yang, Ma, Sun, Wen, Liu, Cai, Ma, et~al.]{zhu2024multi}
Zichen Zhu, Yang Xu, Lu Chen, Jingkai Yang, Yichuan Ma, Yiming Sun, Hailin Wen, Jiaqi Liu, Jinyu Cai, Yingzi Ma, et~al.
\newblock Multi: Multimodal understanding leaderboard with text and images.
\newblock \emph{arXiv preprint arXiv:2402.03173}, 2024{\natexlab{c}}.

\end{thebibliography}
}

\end{document}